\pdfoutput=1

\documentclass[11pt]{article}
\newenvironment{teaserfigure}[1][htbp]
  {\begin{figure*}[#1]\centering}
  {\end{figure*}}
\usepackage[final]{acl}

\usepackage{times}
\usepackage{latexsym}

\usepackage[T1]{fontenc}

\usepackage[utf8]{inputenc}

\usepackage{microtype}

\usepackage{inconsolata}

\usepackage{graphicx}
\usepackage{listings}

\usepackage{xcolor}
\usepackage{xspace}
\usepackage{supertabular}
\usepackage{longtable}
\usepackage{tabularx}
\usepackage{balance}
\usepackage{booktabs}
\usepackage{tcolorbox}
\usepackage{enumitem}  
\usepackage{multirow}
\usepackage{multicol}
\usepackage{makecell}
\usepackage{amsmath}
\usepackage{enumerate}

\lstdefinelanguage{json}{
    basicstyle=\normalfont\ttfamily,
    commentstyle=\color{eclipseStrings}, 
    stringstyle=\color{eclipseKeywords}, 
    numbers=left,
    numberstyle=\scriptsize,
    stepnumber=1,
    numbersep=6pt,
    showstringspaces=false,
    breaklines=true,
    frame=lines,
    string=[s]{"}{"},
    comment=[l]{:\ "},
    morecomment=[l]{:"},
    literate=
        *{0}{{{\color{numb}0}}}{1}
         {1}{{{\color{numb}1}}}{1}
         {2}{{{\color{numb}2}}}{1}
         {3}{{{\color{numb}3}}}{1}
         {4}{{{\color{numb}4}}}{1}
         {5}{{{\color{numb}5}}}{1}
         {6}{{{\color{numb}6}}}{1}
         {7}{{{\color{numb}7}}}{1}
         {8}{{{\color{numb}8}}}{1}
         {9}{{{\color{numb}9}}}{1}
}

\lstdefinelanguage{yaml}{
  keywords={true,false,null,y,n},
  sensitive=false,
  breaklines=true,
  frame=lines,
  comment=[l]{\#},
  morecomment=[s]{/*}{*/},
  morestring=[b]',
  morestring=[b]",
  showstringspaces=false,
  commentstyle=\color{gray},
  keywordstyle=\color{orange},
  basicstyle=\ttfamily\small,
  stringstyle=\color{blue}, 
  moredelim=[s][\color{blue}]{"}{"}, 
  moredelim=[is][\color{black}]{'}{'}, 
}

\lstdefinestyle{yaml2}{
  language=yaml,
  frame=none 
}

\newcommand{\ourmethod}{{AXIS}\xspace}

\title{AXIS: Efficient Human-Agent-Computer Interaction with \\ API-First LLM-Based Agents}

\author{
  \textbf{Junting Lu\textsuperscript{1*}},
  \textbf{Zhiyang Zhang\textsuperscript{2}\thanks{Equal Contribution. Work is done during the internship at Microsoft.}},
  \textbf{Fangkai Yang\textsuperscript{3}\thanks{Corresponding author.}},
  \textbf{Jue Zhang\textsuperscript{3}},
  \textbf{Lu Wang\textsuperscript{3}},
\\ 
  \textbf{Chao Du\textsuperscript{3},}
  \textbf{Qingwei Lin\textsuperscript{3},}
  \textbf{Saravan Rajmohan\textsuperscript{3},}
  \textbf{Dongmei Zhang\textsuperscript{3},}
  \textbf{Qi Zhang\textsuperscript{3}}
\\
  \textsuperscript{1}Peking University,
  \textsuperscript{2}Nanjing University,
  \textsuperscript{3}Microsoft
\\
  aidan.lew.37@stu.pku.edu.cn
}

\begin{document}

\maketitle

\begin{teaserfigure}
  \centering
  \includegraphics[width=\textwidth]{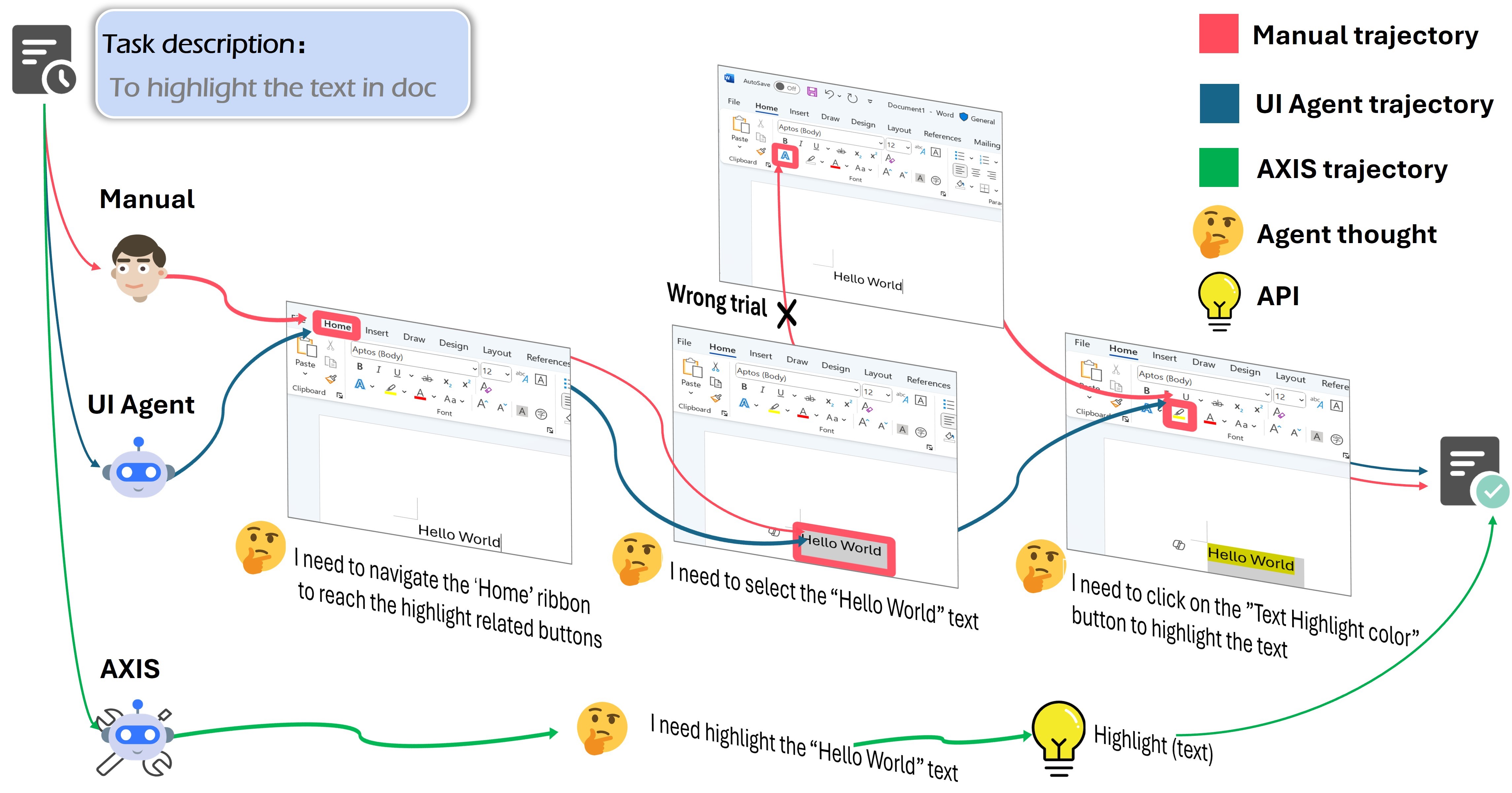}
  \caption{
  An illustration comparing task completion methods: manual operation, UI Agent, and our approach \ourmethod. Manual operation risks wrong trails if users are unfamiliar with the UI. The UI Agent requires numerous sequential interactions. Our \ourmethod efficiently completes the task with a single API call.
  }
  \label{fig:teaser}
  \vspace{-3mm}
\end{teaserfigure}

\begin{abstract}
Multimodal large language models (MLLMs) have enabled LLM-based agents to directly interact with application user interfaces (UIs), enhancing agents' performance in complex tasks. 
However, these agents often suffer from high latency and low reliability due to the extensive sequential UI interactions.
To address this issue, we propose \ourmethod, a novel LLM-based agents framework that prioritize actions through application programming interfaces (APIs) over UI actions. This framework also facilitates the creation and expansion of APIs through automated exploration of applications. Our experiments on Microsoft Word demonstrate that AXIS reduces task completion time by 65\%-70\% and cognitive workload by 38\%-53\%, while maintaining accuracy of 97\%-98\% compared to humans. Our work contributes to a new human-agent-computer interaction (HACI) framework and explores a fresh UI design principle for application providers to turn applications into agents in the era of LLMs, paving the way towards an agent-centric operating system (Agent OS).


\end{abstract}

\section{Introduction}


As personal computers and mobile devices become indispensable in daily life, application industries face pressure to rapidly evolve software with new features to meet growing demands~\cite{ruparelia2010software,abrahamsson2017agile}. However, 
to use a new application effectively, users must first spend time familiarizing themselves with the user interface (UI) and its functionalities, increasing users' cognitive burden~\cite{van2005cognitive,biswas2005learning,plass2010cognitive,darejeh2022cognitive}.
Large language models (LLMs)~\cite{ouyang2022training,achiam2023gpt,dubey2024llama} have demonstrated near-human capabilities in reasoning, planning, and collaboration and are highly promising in completing complex tasks ~\cite{huang2022towards,wei2022chain,mandi2024roco}. Since then, researchers have been leveraging  
multimodal large language models (MLLMs)~\cite{yin2023survey,durante2024agent,zhang2024mm} to operate software applications with 
vision capability~\cite{wu2023can,zheng2024gpt}. Recent works~\cite{yang2023appagent,wang2024mobile,zhang2024ufo,zheng2024gpt} utilize MLLMs to design LLM-based UI agents capable of serving as user delegates, translating user requests expressed in natural language, and directly interacting with the application's UIs to fulfill users' needs without a deep understanding of UIs and functionalities.

However, just like the transition from steam-powered to electric-powered industry took much more than replacing central steam engines with electric motors in the factories, simply building LLM-based UI agent cannot magically deliver a satisfying and worry-free user experience. In particular, today's application UIs are designed for human-computer interaction (HCI)~\cite{lewis1998designing,bradshaw2017human}, which often involves multiple UI interactions for completing a single task. For instance, inserting a 2$\times$2 table in an Microsoft Word document requires a sequence of UI interactions: ``Insert $\rightarrow$ Table $\rightarrow$ 2$\times$2 Table''. Although the HCI-based design suits the habits of humans, training LLM-based UI agents to emulate such interactions would pose quite a few challenges that are difficult to overcome.

The first challenge for the LLM-based UI agent is high latency and long response time. Each individual UI interaction step requires one LLM call to reason which UI to interact with. A task involves multiple UI interaction steps can thus incur considerable time and monetary costs. The LLM call latency is also positively correlated with the number of processed tokens~\cite{levy2024same,wang2024model,egiazarian2024extreme}. To ensure the LLM returns high-quality outputs, the LLM-based UI agent must pass a large volume of UI information to precisely describe the current state, which also increases latency in each call.

The second challenge lies in the domain of reliability. Studies have shown that LLMs are prone to hallucinations in generating responses and selecting correct UIs when a long chain of UI interactions is required~\cite{bang2023multitask,dhuliawala2023chain,zhang2023siren,guan2024intelligent}. During long sequential calls, the chance of selecting a wrong UI control or hallucinating a non-existent UI for interaction increases with each reasoning step~\cite{zhang2024ufo}, resulting in compounding errors and task failure~\cite{chen2024more,zhao2024large}. 
Lastly, the LLM-based UI agent also faces the challenge of UI generalization. While recent works had made advancements in UI grounding~\cite{cheng2024seeclick,rawles2024androidinthewild,Qwen-VL}, how the LLM-based UI agents handle interactions with applications whose UIs are not included in the pretraining stage of LLMs remains a critical obstacle without good solutions. 

To address these challenges faced by LLM-based UI agents, we highlight the necessity of application programming interfaces (APIs) within the new human-agent-computer interaction (HACI) paradigm. As an intermediate layer between human and computer, the LLM-based agent should understand user requests in natural language and operate computers/applications by prioritizing APIs over human-like UI interactions. Inspired by this, we propose \textbf{\ourmethod}: \textbf{A}gent e\textbf{X}ploring AP\textbf{I} for \textbf{S}kill integration, a self-exploration LLM-based framework capable of automatically exploring existing applications, learning insights from the support documents and action trajectories, and constructing new APIs\footnote{The new APIs are also referred as ``Skills'' in Section~\ref{sec:method_design}, and we use the term ``API'' loosely here to differentiate from the UI.} based on the existing ones. \ourmethod helps build API-first LLM-based agents that replace UI agents to prioritize API calls over unnecessary multistep UI interactions in task completion. Regular UI interactions are only called when the related APIs are unavailable. Figure~\ref{fig:teaser} shows a task completion with the UI agent and API-first agent. Compared to the UI agents, API-first agents require fewer tokens and can obtain more accurate, code-formatted responses from LLMs with low latency and high reliability. 

Our work makes the following contributions:

\begin{itemize}[nosep, leftmargin=*]
    \item We propose a novel HACI paradigm along with an implementation framework, \ourmethod, which enables API-first LLM-based agents to explore an application’s available APIs and construct new ones as needed. This approach facilitates the straightforward transformation of any application into an autonomous agent by wrapping it with an API set and adopting a simplified UI design. This paradigm not only addresses practical challenges faced by application providers but also paves the way toward a full-fledged Agent Operating System (Agent OS)~\cite{zhang2024vision,mei2024aios,wu2024copilot}.
    
    \item We mitigate cognitive load and reduce the learning effort required for complex interactions by replacing multi-step UI operations with efficient API calls. Our experiments on Microsoft Word tasks~\cite{officeword} show that \ourmethod significantly improves task completion rates while alleviating the cognitive burden on users.
    
    \item We conduct comprehensive performance evaluations and an extensive user study to validate the efficiency, reliability, and practical applicability of \ourmethod in real-world scenarios.
\end{itemize}

\section{Related Work}

\subsection{LLM-based UI Agent}\label{sec:llm-basedguiagent}
LLM-based agents are designed to utilize the advanced context understanding and reasoning skills of LLMs to interact with and manipulate environments with human-like cognitive abilities~\cite{meta2022human, xi2023rise, liu2023agentbench, wang2024survey}. 
The advent of MLLMs~\cite{yin2023survey, durante2024agent, zhang2024mm}, including GPT-4o~\cite{GPT-4o} and Gemini~\cite{team2023gemini}, expands the research landscape for LLM-based UI agents. LLM-based UI agents have been applied to multiple areas such as mobile platforms~\cite{yan2023gpt, yang2023appagent, wang2024mobile}, Web~\cite{song2024beyond} and OS ~\cite{wang2024large, zhang2024ufo, zheng2024gpt, tan2024towards, hong2024cogagent, cheng2024seeclick}.
LLM-based UI agents mimic human interactions, but existing UIs are designed for human-computer, not agent-computer, interactions, leading to inefficiencies in repetitive tasks. APIs offer a more efficient alternative by reducing unnecessary UI steps. We explore leveraging APIs for LLM-based agents and propose new UI design principles for the LLM era.

\subsection{Agent Operating System}
To support the completion of complex tasks with minimal human interventions, emerging works have explored the possibility of developing an agent operating system (Agent OS) fully supported by LLMs~\cite{mei2024aios,wu2024copilot,zhang2024vision,xie2024osworld,rawles2024androidworld}. In the industry, commercial Agent OSes~\cite{appleintelligence,copilotPC,harmonyOS,magicOS} are evolving to become more accessible and productive for customers with the potential of leading a new era of HCI. 
Existing Agent OSes typically divide complex tasks into sub-tasks and assign them to applications. However, LLM-based agents still rely on human-like UI interactions, such as clicking and swiping, which are less efficient than API calls. Additionally, when processing a task, the LLM-based UI takes control away from the user.

\subsection{UI design in LLM era}
UI design is an essential part of HCI and requires highly specialized expertise along with iterative rounds of feedback and revision~\cite{stone2005user}. With LLMs, UI design can be further empowered with automated procedures of design, feedback and evaluation.~\citet{duan2024generating} use LLM-generated feedback to automatically evaluate UI mockups. Similiarly, SimUser~\cite{xiang2024simuser} leverages LLMs to simulate users with different characteristics to generate feedback on usability and provide insights into UI design. MUD~\cite{feng2024mud} utilize LLMs to mimic human-like exploration to mine UI data from applications and employs noise filtering to improve quality of UI data.
Existing UI designs follow the traditional HCI paradigm rather than the HACI paradigm central to Agent OS. Using the \ourmethod framework, we explore applications, identify essential UIs, and determine which UI components can be replaced by API calls for LLM-based agents.

\section{Preliminary}


\noindent\textbf{Environment.}
In the context of \ourmethod, the environment refers to the collection of interactive entities within the exploration scope of the agents. In our case, these entities primarily consist of applications running on the Windows operating system, with Microsoft Word being our focus. Applications in the environment often share common elements, such as controls ~\cite{zhang2024ufo} and XML elements obtained after unpacking. To facilitate the observation and interaction between agents and the environment, we have designed two general interfaces: \texttt{state()} to return the environment state and \texttt{step()} to execute agent actions, respectively. Details of interfaces are in Appendix~\ref{appendix:interfaces}.

\noindent\textbf{Skills.}
A skill in \ourmethod is a structured unit designed to accomplish a specific task within the environment. It is a high-level representation of UI- and API-based actions, with priority given to API actions\footnote{If the skill can be represented with UI or API actions, the skill is represented in API-only actions.}. Following the design of tool usage and function call~\cite{cai2023large, wang2023voyageropenendedembodiedagent}, each skill consists of three components: description, skill code, and usage example. Appendix~\ref{appendix:skillcomponent} shows details of each component.

\noindent\textbf{Skill Types and Hierarchy.} Following a versatile design principle, the skills in \ourmethod can be categorized into five types based on the composition of their code fragments: Atomic UI Skill, Atomic API Skill, Composite UI Skill, Composite API Skill, and API-UI Hybrid Skill, the details are shown in Appendix~\ref{appendix:skilltype} Table~\ref{tab:type}. Additionally, we define ``skill hierarchy'' as the number of skills contained. A single atomic skill thus has a skill hierarchy of~\textit{1}. Based on their code composition and hierarchy, skills in \ourmethod can be nested. For example, skill A can call skill B, which in turn calls skill C, forming a hierarchy of depth~\textit{3}.

\section{Design of \ourmethod}\label{sec:method_design}
\begin{figure*}[htbp]
  \centering
  \includegraphics[width=\textwidth]{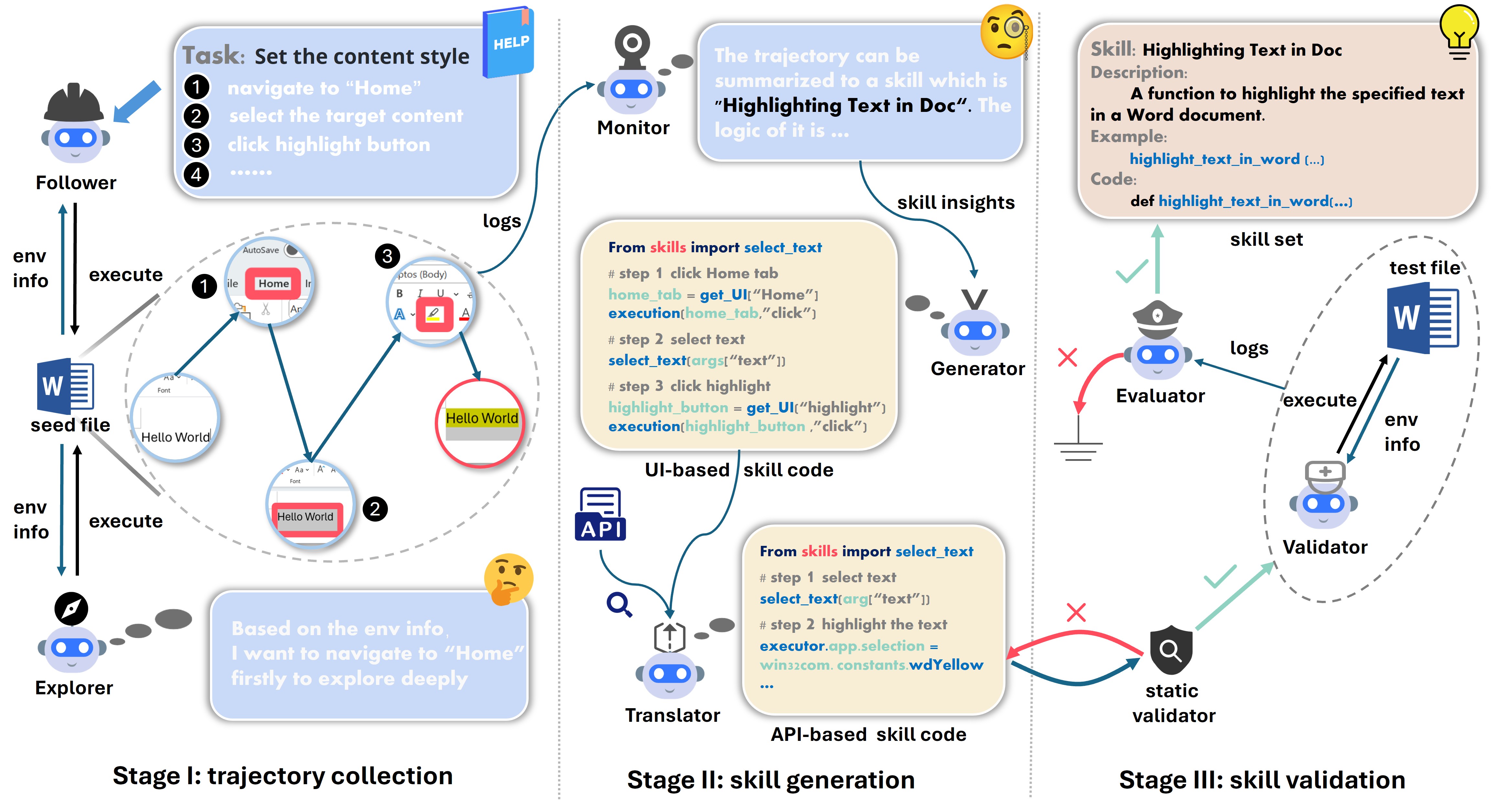}
  \caption{Overview of \ourmethod framework. AXIS first collects interaction trajectories in Follower or Explorer mode. Then, the explored trajectories are used to generate skills and translate them into skill code. The skill validation stage then validates skills in the real environment. Note that the dashed boxes refer to the interaction between agents and application environment.}
  \label{fig:general-framework}
\end{figure*}

We develop \ourmethod as a framework that can automatically explore within existing application environments, learn insights from exploration trajectories, and consolidate available insights and learned knowledge into actionable ``skills''. Illustrated by Figure~\ref{fig:general-framework}, the \ourmethod system is composed of three crucial stages: \textbf{trajectory collection}, \textbf{skill generation}, and \textbf{skill validation}. The trajectory collection stage collects interaction trajectories in task completion, and then the skill generation stage generates skills from these trajectories and translates into skill code. The skill validation stage validates the skill code before it is added in the skill library to reduce hallucination and maintain generalizability.


\subsection{Stage I: Trajectory Collection}

\textit{``Experience is the mother of wisdom''}. Following the approach in \cite{wang2023voyageropenendedembodiedagent}, we let the agent practice acquiring skills in the application environment. In this stage, we designed two practice modes: \textit{Follower Mode} and \textit{Explorer Mode}, which are task-oriented collections with and without task descriptions, respectively. The specific prompts refer to Appendix~\ref{sec:traj-prompts}. 


\noindent\textbf{Follower Mode.} The Follower Agent extracts tasks and instructions from the application help documents. The agent processes structured inputs: task requirements, step-by-step instructions, environmental context, and the current skill library. It strictly adheres to instructions while interacting with seed application files, collecting execution trajectories. Action selection employs the ReAct mechanism~\cite{yang2023dawn}, integrating action history, step descriptions, and environmental states. Initially limited to basic actions (see Table~\ref{tab:basic}), the follower agent is able to perform more complex tasks with the expansion of the skill library through the other parallel processing stages.

\noindent\textbf{Explorer Mode.} Explorer mode differs from the follower mode by leveraging the brainstorming capability of LLMs to generate various step-by-step instructions. The Explorer agent
observes environment states and action history, selecting actions from the skill library without following predefined task guides.
To enhance state diversity, we implemented:
\begin{itemize}[nosep, leftmargin=*]
\item \textit{Initial State Diversity}: Random replay of Microsoft Word files up to intermediate steps, providing varied starting points.
\item \textit{Vertical/Horizontal Exploration}: A random ``dive into'' strategy balancing continuity (vertical) and breadth (horizontal) in functionality exploration. ``Vertical'' means the agent will select UI controls at lower levels of the current control tree for its next action, while ``Horizontal'' means the agent will tend to select UI controls at the same level or switch to other menus.
\item \textit{Skill Proficiency Levels}: Three explorer levels (low/medium/high) guided by Microsoft Office Specialist (MOS) certification curriculum\footnote{https://learn.microsoft.com/en-us/credentials/certifications/mos-word-2019/}, with system prompts tailored to their app familiarity.
\end{itemize}

\subsection{Stage II: Skill Generation}
This stage runs in parallel with Stage I and is responsible for converting the trajectories into structured skills. This process involves three LLM-based agents. The specific prompts for these agents are detailed in Appendix~\ref{sec:skillgen-prompts}.

\noindent\textbf{Monitor.} 
The collected trajectories often contain complex and multiple interaction steps, making it challenging to convert them into a single skill without enough basic skills available in the skill library. The monitor examines the skill library, extracts meaningful trajectory segments, and consolidates them into natural language-based skill insights.  

\noindent\textbf{Generator.} The Generator agent faithfully combines the skill insights with their corresponding trajectory segments and generates executable skill code.
As shown in Figure~\ref{fig:general-framework}, the skill code mainly consists of stacked actions from the trajectories, where the original action parameters are converted into placeholders. Along with the skill code generation, it also produces corresponding skill usage examples and descriptions, with examples shown in Table~\ref{tab:skill-trans}. Note that since the skill code is a faithful reproduction of the trajectory, which may contain numerous UI operations, the skill code essentially becomes a combination of UI actions, leaving room for speed optimization.

\noindent\textbf{Translator.} As the skill code from the generator contains numerous UI operations, a ``UI'' to ``API'' translation is needed. The Translator agent accomplishes this task by connecting to the retrieval-augmented generation (RAG) module~\cite{gao2023retrieval}, consulting the application's documentation and the current skill library to convert UI operation code segments to APIs. As shown in Figure~\ref{fig:general-framework}, it converts UI operations into direct modifications of target content within the application. Table~\ref{tab:skill-trans} shows a translation example.

\subsection{Stage III: Skill Validation}
Due to potential LLM hallucinations and cumulative errors in multi-agent transmission, generated skills may have syntax or functionality issues. To address this, we implement static and dynamic validation, with dynamic validation following static validation.

\noindent\textbf{Static Validation.} Static validation utilizes structural checks to verify the compatibility of the skill code with the skill executor. It examines whether the skill's parameters contain the mandatory parameters (such as the executor instance and args list), whether the methods and properties of the executor are correctly invoked in the code, and whether any non-existent skills are imported when reusing the skill. Skills that do not meet these formal requirements are returned for revision.

\noindent\textbf{Dynamic Validation.} Dynamic validation evaluates the performance of a skill in a real environment. It consists of a validator and an evaluator. The validator generates various input parameters to test the skill's generalizability, and the evaluator checks whether the test is successfully completed by examining the final state. The specific prompts are detailed in Appendix~\ref{sec:skillval-prompts}, with detailed description of the validator and evaluator in Appendix~\ref{appendix:dynamicvalidator}.

\section{Feasibility Study}
To validate the usability and effectiveness of the AXIS framework, we conduct a feasibility study. We first use AXIS to explore Microsoft Word and discover 73 skills. Then we extract 50 tasks from the wikihow \footnote{https://www.wikihow.com/Use-Microsoft-Word} page ``Use Microsoft Word'' and the official Microsoft Word website \footnote{https://support.microsoft.com/en-us/word}.
These tasks were executed using both AXIS and UI Agent, and the results were analyzed and compared. AXIS also enable turn an application into an agent by simply wrapping the application with an API set and adopting a simpler UI design suitable for HACI, which is presented in Appendix~\ref{sec:extensive}.
\subsection{SKill Exploration}
Before the exploration, AXIS is provided with the initial skill library which composed of 6 basic actions as shown in Table~\ref{tab:basic}. Then, 347 seed files are used for AXIS to explore. After the exploration, AXIS discovers 73 skills with different hierarchies. Majority of the skills (44) discovered have a skill hierarchy 1. The rest is composed of 24 skills with hierarchy 2, 3 skills with hierarchy 3, and 2 skills with hierarchy 4. Table~\ref{tab:explored} displays several successfully validated skills discovered during the exploration process. 

\subsection{Task Completion}

We evaluate UI Agent (represented by UFO~\cite{zhang2024ufo} for its superior Word task performance) and AXIS on 50 Word-related tasks using explored skills. Table~\ref{tab:res50} presents comparative results, including completion time, success rates, step counts, and LLM backend costs (GPT-4o, version 20240513) for both agents.

In terms of execution time, AXIS demonstrates superior performance, completing tasks in 29.9 seconds on average - twice as fast as UI Agent's 59.5 seconds. It also achieves higher success rates and, through its skill-based abstraction, requires fewer steps per task, resulting in lower costs compared to the UI Agent UFO.


\begin{table}[!ht]
    \centering

    \setlength{\tabcolsep}{2pt} 
    \small
    \begin{tabular}{lccl}
    \hline
        \textbf{Metric} & \textbf{UI Agent} & \textbf{AXIS} & \textbf{P - Significance}\\ \hline
        Time(s) & 59.5 & 29.9 & u>a (p < 0.001)\\ 
        Success Rate(\%) & 52.0 & 84.0 & u<a (p < 0.001)\\ 
        Steps & 3.2 & 2.0 & u>a (p < 0.01)\\ 
        Cost(\$)  & 0.4 & 0.2 & u>a (p < 0.001)\\ \hline
    \end{tabular}
    \caption{Comparison of the performance of UI Agent and AXIS on 50 tasks. Here, “P - Significance” represents “Pairwise Significance”.}
    \label{tab:res50}
\end{table}

To investigate AXIS's efficiency, we analyze action types and API usage patterns. As shown in Table~\ref{tab:resactions50}, AXIS employs significantly fewer UI actions compared to the UI Agent, while utilizing more API and Advanced API calls (skills with hierarchy level $\geq$ 2).
Analysis reveals AXIS predominantly employs integrated API skills for task completion, yielding more API actions (55.7\% usage rate, including 23.1\% advanced APIs) compared to UI Agent's 8.1\%. This demonstrates AXIS's API-first approach, leveraging available skills for efficient task execution through skill-action integration.

\begin{table}[t]
    \centering
    \small
    \begin{tabular}{lccl}
    \hline
        \textbf{Metric} & \textbf{UI Agent} & \textbf{AXIS} \\ \hline
        Total UI actions & 103 & 48 \\ 
        Total API actions & 9 & 39 \\ 
        API usage rate(\%) & 8.1 & 55.7  \\ 
        Advanced API usage rate(\%)  & - & 23.1 \\ \hline
    \end{tabular}
    \caption{Comparison of hit UI actions and API actions of UI Agent and AXIS on 50 tasks.}
    \label{tab:resactions50}
    \vspace{-3mm}
\end{table}

\section{User Study}

We conduct an extensive user experiment to evaluate the performance of \ourmethod. The experiment and evaluation metrics are designed to explore the following research questions (RQs) on the role of LLM-based agents in work and daily life scenarios:
\begin{itemize}[nosep, leftmargin=*]
    \item \textbf{RQ1}: Does the LLM-based agent lower the cognitive load of the users and make them have less effort to learn?
    \item \textbf{RQ2}: Does the LLM-based agent enhance the efficiency of users? 
    \item \textbf{RQ3}: What are the differences between a UI Agent and an API-based Agent in user experience? 
\end{itemize}

In our user experiment, participants are asked to complete specified tasks within an application through three methods: manually, with the assistance of a UI Agent, and with the assistance of AXIS. The entire process is recorded. Microsoft Word is chosen as the experimental application considering  its popularity in our daily work and life as well as the rich API documentations~\cite{wordapi}). Thus five tasks of Word are sampled from both official Word documentation and GPT-generated results, divided into two difficulty levels: low difficulty (L1) and high difficulty (L2). Motivated by the RQs, we set three objectives for the user experiment:
(1) To evaluate the cognitive load on participants when completing tasks using different methods.
(2) To compare the efficiency and reliability of task completion across the three methods.
(3) To assess user preferences regarding the use of different Agents. This study is approved by the Institutional Review Board (IRB) of University.

The details of experiment procedure and participants recruitment are shown in Appendix~\ref{sec:user-study}.

\begin{table*}[!ht]
    \centering
    \small
    \begin{tabular}{lclccc}
    \hline
        \textbf{Metric} & \textbf{Task Level} & \textbf{Manual} & \textbf{UI Agent} & \textbf{AXIS} & \textbf{Pairwise Significance} \\ \hline
        Time(s) & \makecell{L1 \\ L2} & \makecell{61.8 \\ 167.6} & \makecell{104.6 \\ 155.5} & \makecell{18.2 \\ 57.1} & \makecell[r]{\textbf{L1}: m<u (p < 0.001) \\ \textbf{L1, L2}: a<m (p < 0.001) \\ \textbf{L1, L2}: a<u (p < 0.001)} \\ \hline
        Success Rate(\%) & \makecell{L1 \\ L2} & \makecell{100.0 \\ 97.5} & \makecell{75.0 \\ 45.0} & \makecell{98.3 \\ 95.0} & \makecell[r]{\textbf{L1, L2}: m>u (p < 0.001) \\ \textbf{L1, L2}: a>u (p < 0.001)} \\ \hline
    \end{tabular}
    \caption{Comparison of Methods on Time and Success Rate in L1 and L2 tasks.}
    \label{tab:time}
\end{table*}

\begin{table*}[!ht]
    \centering
    \small
    \begin{tabular}{lcccc}
    \hline
        \textbf{Metric} & \textbf{Task Level} & \textbf{UI Agent} & \textbf{AXIS} & \textbf{Pairwise Significance} \\ \hline
        steps & \makecell[l]{L1\\L2} & \makecell[l]{6.4\\11.1} &  \makecell[l]{1.0\\4.2} & \makecell[l]{\textbf{L1}: a<u (p < 0.001)\\\textbf{L2}: a<u (p < 0.001)}
        \\ \hline
        cost(\$) & \makecell[l]{L1\\L2} & \makecell[l]{0.6\\0.9}  & \makecell[l]{0.07\\0.3}  &  \makecell[l]{\textbf{L1}: a<u (p < 0.001)\\\textbf{L2}: a<u (p < 0.001)} 
        \\ \hline
    \end{tabular}
    \caption{Comparison of Methods on Steps and Cost in L1 and L2 tasks.}
    \label{tab:steps}
\end{table*}

\subsection{Experimental Metrics} We collect both subjective and objective metrics in our experiments to evaluate the performance and user experience of different methods.

\noindent\textbf{Subjective Metrics.} 
As detailed in Appendix~\ref{sec:user-study}, we conduct four post-task questionnaires (manual and Agent-assisted L1/L2 tasks) using NASA-TLX~\cite{hart1988development} metrics: Mental, Physical, and Temporal Demand; Performance; Frustration; and Effort, supplemented by a learning effort metric. Lower scores indicate reduced cognitive load, improved success perception, and decreased frustration/effort. For Agent-assisted tasks (Questionnaires 3-4), additional metrics include Agent fluency/reliability, UI dependency, decision consistency (Agent-user alignment), and perceived completion speed.


\noindent\textbf{Objective Metrics.}
For objective metrics, we maintain comprehensive experimental logs, including screen recordings of manual and Agent-assisted tasks, decision-making processes, UI interaction paths, task completion time and success rates, LLM backend costs (GPT-4, version 20240513) for Agent operations, and UI dependency measurements for both the UI Agent and AXIS.

\begin{table*}[!t]
    \centering
    \renewcommand{\arraystretch}{1.5}
    \small
    \begin{tabular}{lcclc}
    \hline
       \textbf{Metric} & \textbf{Task Level} & \textbf{Manual} & \textbf{Agents} & \textbf{Pairwise Significance} \\ 
        \hline
        Mental Demand (0-100) & \makecell{L1 \\ L2} & \makecell{21.3 \\ 70.0} & \makecell{2.5 \\ 7.5} & \makecell[l]{\textbf{L1}: m>a (p < 0.001) \\ \textbf{L2}: m>a (p < 0.001)} \\ 
        \hline
        Physical Demand (0-100) & \makecell{L1 \\ L2} & \makecell{31.3 \\ 57.5} & \makecell{5.0 \\ 6.3} & \makecell[l]{\textbf{L1}: m>a (p < 0.001) \\ \textbf{L2}: m>a (p < 0.001)} \\ 
        \hline
        Temporal Demand (0-100) & \makecell{L1 \\ L2} & \makecell{52.5 \\ 37.5} & \makecell{28.8 \\ 35.0} & \makecell[l]{\textbf{L1}: m>a (p < 0.05) \\ \textbf{L2}: -} \\ 
        \hline
        Performance (0-100) & \makecell{L1 \\ L2} & \makecell{21.2 \\ 47.5} & \makecell{21.2 \\ 26.2} & \makecell[l]{\textbf{L1}: - \\ \textbf{L2}: m>a (p < 0.05)} \\ 
        \hline
        Frustration Level (0-100) & \makecell{L1 \\ L2} & \makecell{31.3 \\ 62.5} & \makecell{7.5 \\ 10.0} & \makecell[l]{\textbf{L1}: m>a (p < 0.001) \\ \textbf{L2}: m>a (p < 0.001)} \\ 
        \hline
        Completion Effort (0-100) & \makecell{L1 \\ L2} & \makecell{12.5 \\ 35.0} & \makecell{17.5 \\ 13.8} & \makecell[l]{\textbf{L1}: - \\ \textbf{L2}: m>a (p < 0.01)} \\ 
        \hline
    \end{tabular}
    \caption{Comparison of NASA-TLX results of Level 1 and Level 2 tasks. (m: Manual, a: Agents)}
    \label{tab:nasa-tlx}
\end{table*}

\subsection{Results}
Our analysis explores three key aspects: (1) cognitive load reduction, (2) task efficiency of agents, and (3) user preferences between UI agents and AXIS.

\begin{figure*}[!ht]
    \centering
    \includegraphics[width=\textwidth]{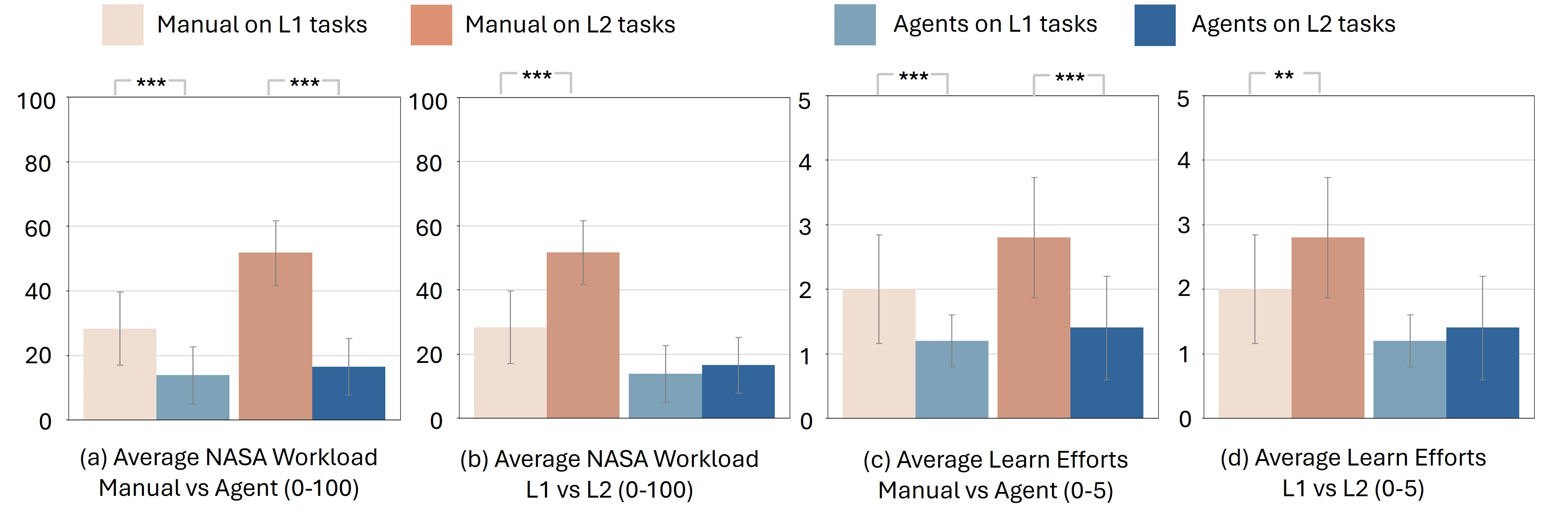}
    \caption{The results of NASA Workload and learn efforts on L1 and L2 tasks of user study. Bars indicate standard errors (**: p < 0.01, ***: p < 0.001)}
    \label{fig:cognitives}
\end{figure*}

\noindent\textbf{Cognitive Load.} To assess cognitive load reduction by LLM-based Agents, we analyze NASA-TLX and learning effort scores (Table~\ref{tab:nasa-tlx}, Figure~\ref{fig:cognitives}). Results show higher scores for L2 tasks across dimensions, validating our task difficulty classification. Agent-based methods significantly outperform manual approaches in reducing mental/physical demand and frustration, particularly for complex L2 tasks (p<0.05). While L1 task performance differences were minimal, agents notably enhanced users' success perception in L2 tasks. Analysis reveals consistent user experiences across task complexities (Figure~\ref{fig:cognitives} (b)) and significant learning effort reduction, especially for difficult tasks (Figure~\ref{fig:cognitives} (c)). These findings answer \textbf{RQ1}: LLM-based agents effectively reduce cognitive load and learning effort, particularly for complex tasks.


\begin{figure*}[!ht]
    \centering
    \includegraphics[width=\textwidth]{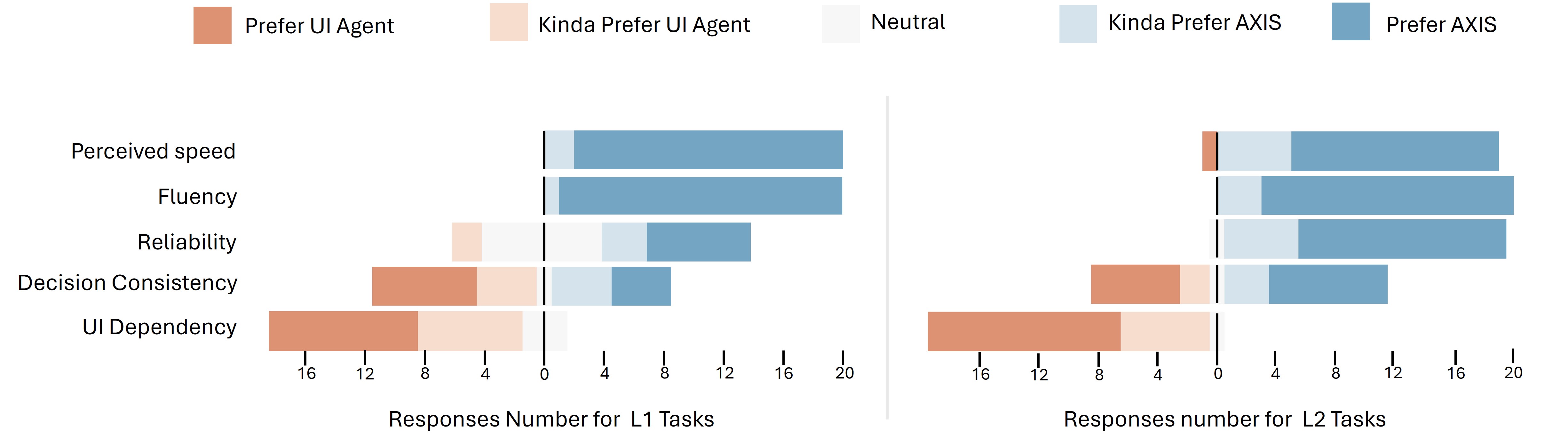}
    \caption{The results of subjective preference on L1 and L2 tasks of user study.}
    \label{fig:preferences}
\end{figure*}

\noindent\textbf{Efficiency and Reliability.} 
Our evaluation compares manual, UI Agent, and \ourmethod approaches using completion time, success rate, steps, and costs (summarized in Tables~\ref{tab:time} and~\ref{tab:steps}). AXIS demonstrates superior time efficiency, significantly outperforming both methods (p < 0.001), particularly in complex L2 tasks. While manual methods achieve highest accuracy, AXIS maintain near-human performance, contrasting with UI Agent's lower accuracy due to UI element positioning and visibility issues. Notably, UI Agents require substantially more steps for deeply nested L2 tasks, whereas AXIS's API-driven approach enable streamlined execution with fewer steps and lower costs. This answers \textbf{RQ2}: while UI Agents offer modest efficiency gains for complex tasks with reliability challenges, AXIS consistently improves human efficiency with superior reliability.

\noindent\textbf{Affinity Preference.} To compare user experiences between UI agents and AXIS, we conduct a subjective evaluation across five aspects (Figure~\ref{fig:preferences}). Participants consistently preferr AXIS for its perceived speed, fluency, and reliability across both L1 and L2 tasks. While AXIS's highly encapsulated API led to less human-like decision-making in simple tasks (L1), its decisions aligned better with human reasoning in complex tasks (L2). AXIS notably improves user's experienct by reducing UI dependency compared to UI agents' frequent interface interactions. These findings answer \textbf{RQ3}: AXIS provides superior efficiency, smoothness, and reliability compared to UI agents, with increasing user preference as task complexity grows. User feedback highlights AXIS's API-first approach as advantageous over UI agents' intrusive behaviors, offering better control and experience.

\section{Conclusion}
We present AXIS, a framework that enhances human-agent-computer interaction (HACI) by prioritizing API calls over UI interactions to reduce inefficiencies and cognitive burdens in complex tasks with multimodal large language models (MLLMs). Experiments with Microsoft Word show that AXIS cuts task completion time by 65\%-70\% and cognitive workload by 38\%-53\%, while maintaining human-level accuracy. These results demonstrate the potential of API-first LLM-based agents to streamline interactions, reduce latency, and improve reliability, offering a novel approach to faster, more efficient task execution.

\section{Ethical Statement}
All used datasets collected internally or obtained from external sources, ensuring no infringement on individual or organizational rights. User study participants volunteered and were compensated.
\newpage
\section{Limitations}
While AXIS effectively mines APIs and enables efficient human-agent-computer interaction (HACI), further optimization is needed to achieve an Agent OS. Currently, AXIS primarily relies on Python-based APIs, limiting support for applications without native Python interfaces. Additionally, its exploration process requires improvements in stability and efficiency. Future work should focus on developing unified action interfaces to extend HACI to more applications and operating systems while enhancing the framework’s performance and efficiency.

\bibliography{main}

\appendix

\section{Interfaces in Environment}\label{appendix:interfaces}
Below are the details of two interfaces: \texttt{state()}
and \texttt{step()} to return the environment state and execute agent actions, respectively. 

\begin{itemize}
    \item \texttt{state()}: Returns the state of the environment, including detailed information on the current elements within the environment. The environment state encompasses key UI information such as control positions, control types, and selection status, which is consistent with the definition in ~\cite{zhang2024ufo}. For applications that can be unpacked, the unpacked XML content is also included as part of the state.
    
    \item \texttt{step()}: Incorporates a skill executor that allows agents to perform operations within the environment by executing skills. Upon completion, this interface returns the results of these operations.
\end{itemize}

\section{Skills}

\subsection{Skill Component}\label{appendix:skillcomponent}
Each skill consists of three main components:

\begin{itemize}
    \item Skill Code: A piece of code structured to be compatible with the skill executor. Skill code includes a uniform set of parameters and adheres to the standard PEP 257 documentation. The initial set of skills is generated by restructuring the fundamental APIs from the application provider, and based on these initial skills, \ourmethod can explore and develop additional new skills.
    
    \item Description: A description of the skill's functionality, which assists the LLM in selecting and invoking appropriate skills during task execution.
    
    \item Usage Example: One or more code examples including typical parameters associated with the code and description. These examples help the LLM in correctly formatting parameters when invoking the skill.
\end{itemize}

\noindent\textbf{Skill Executor.} As discussed above, our application environment incorporates a \texttt{step()} interface to facilitate the interaction between agents and the environment. This interface also hosts the skill executor responsible for executing the skill generated or selected by the agents. The skill executor keeps caching of application documents and simultaneously supports multiple functionalities including locating application controls, invoking methods on those controls, and calling application APIs (independent of controls), to enable the UI actions and API actions in the same time and serve as an efficient foundation for skill-driven operations.

\subsection{Skill Type}\label{appendix:skilltype}
The skills in \ourmethod can be categorized into five types based on the types of the code fragments: Atomic UI Skill, Atomic API Skill, Composite UI Skill, Composite API Skill, and API-UI Hybrid Skill. There are some examples in Table~\ref{tab:type}.

\begin{table*}[!ht]
    \centering
    \caption{Comparison of 4 types of skill.}
    \begin{tabular}{lp{6cm}cp{3cm}}
    \hline
        \textbf{Type} & \textbf{Description} & \textbf{Example} & \textbf{Feature coverage}
        \\ \hline
        Atomic UI skill & Composed of one basic UI action. As the most primitive skills, Atomic UI skills are stacked and transformed during the exploration process to form new skills. & \textit{click\_input} & click on different UI controls
        \\ \hline
        Atomic API skill & Composed of one basic API actions. Unlike UI actions that depend on UI controls for execution, API actions can be executed without the need of interacting with any UI elements. & \textit{select\_text} & select text content in the canvas.
        \\ \hline
        Composite UI skill & Composed of multiple atomic UI actions or composite UI actions. Composite UI skill are formed by a simple stacking and combination of UI actions.
        & \textit{search\_for\_help}
        & clicking the search box and then editing text.
        \\ \hline
        Composite API skill & Composed of multiple atomic API actions or composite API actions. This type of skills often represents a higher-level combination of functions.
        & \textit{insert\_header\_footer}
        & insert header and footer with specified contents by API, which is equal to sequential UI actions "Insert->Header->footer edit->Footer->footer edit".
        \\ \hline
        API-UI hybrid skill & Composed of both API actions and UI actions. API-UI hybrid skills sometimes appear as intermediate states during skill exploration and may evolve into pure API actions during the later stage of exploration. 
        & \textit{format\_text\_in\_word}
        & combine \textit{select\_text} and a series of UI actions related to text styling.
        \\ \hline
        
    \end{tabular}
    \label{tab:type}
\end{table*}

\subsection{Initial Skill Repository}
As the foundation for the interaction between the agent and the environment, before exploring and mining skills, we pre-defined some basic actions as the initial skill library shown in Table~\ref{tab:basic}. These actions are derived from UFO~\cite{zhang2024ufo}. 

\begin{table*}[!ht]
    \centering
    \caption{The basic actions constituted the initial skill repository.}
    \begin{tabular}{cp{3cm}p{3cm}}
    \hline
        \textbf{Name} & \textbf{Description} & \textbf{Example}
         \\ \hline
        \textit{set\_edit\_text} & The function to Set the edit text of the control element, can use to input content on Edit type controls. & \textit{set\_edit\_text(executor,
        args\_dict={"control\_id":'119',
        "control\_name":"Edit",
        'text':"hi there"})}
        \\ \hline
        \textit{select\_text} & A function to select the text with the specified text content. & \textit{select\_text(executor,
        args\_dict={"text":"hello"})}
        \\ \hline
        \textit{select\_table} & A function to select the table with the specified number. & \textit{select\_table(executor,
        args\_dict={"number":1})}
        \\ \hline
        \textit{type\_keys} & A function to Type in keys on control item.Used to enter shortcuts and so on. & \textit{type\_keys(executor,
        args\_dict={"control\_id":'119',
        "control\_name":"Edit",
        "text": "{VK\_CONTROL down}", 
        "newline": False})}
        \\ \hline
        \textit{click\_input} & A function to Click the control element.Usually be used to switch to different ribbon,click the buttons in menu. & \textit{click\_input(executor,
        args\_dict={"control\_id":"12",
        "control\_name":"Border",
        'button':"left",'double':False})}
        \\ \hline
        \textit{wheel\_mouse\_input} & A function for Wheel mouse input on the control element. & \textit{wheel\_mouse\_input(executor,
        args\_dict={"control\_id":"12",
        'wheel\_dist':-20})}
        \\ \hline
    \end{tabular}
    \label{tab:basic}
\end{table*}

\subsection{Skill Translation Example}
The following is a specific example regarding the generation and translation of skills in Table~\ref{tab:skill-trans}.
\onecolumn
\begin{table*}[htbp]
    \centering
    \caption{An example of skill generation and translation.}
    \begin{longtable}{p{0.48\textwidth}|p{0.48\textwidth}}
        \hline
        \textbf{Generated Skill by Generator Agent} & \textbf{Translated Skill by Translator Agent} \\
        \hline
        \begin{lstlisting}[style=yaml2]
code: |- 
from select_text import select_text
def hightlight_text(executor, args_dict: dict):
    """
    Highlight the text.
    :param executor: The executor object.
    :param args_dict: The arguments of the highlight text method in a dict
        - text: the text to highlight
        - home_control_id: the id of the Home ribbon.
    :return: The result of the action.True or False.
    """
    try:
        select_text(executor, args_dict)
        home_control_id = args_dict.get("home_control_id", -1)
        home_ribbon_control = executor.get_target_control_by_args(
            control_args={"control_id":home_control_id})
        executor.atomic_execution(
            control=home_ribbon_control,
            method_name="click_input",
            args={
                "button": "left",
                "double": False
            }
        )
        hightlight_button = executor.get_target_control_by_uuid(
            "uuid_control.json", "96d5a516-35af-459b-b1c4-34aced6acb0b")
        executor.atomic_execution(
            control=hightlight_button,
            method_name="click_input",
            args={
                "button": "left",
                "double": False
            }
        )
    except Exception as e:
        print(f"error:{str(e)}")
        return False
    return True

description: |-
A function to highlight the text.
Note: the args_dict should include "home_control_id" field of "Home" ribbon control id
and "text" field of the target text to highlight.

example: |-
highlight_text(executor, args_dict={"home_control_id": 27, "text": "Hello"})

        \end{lstlisting} &
        \begin{lstlisting}[style=yaml2]
code: |- 
from select_text import select_text
def highlight_text(executor, args_dict: dict):
    """
    Highlight the text.
    :param executor: The executor object.
    :param args_dict: The arguments of the highlight text method in a dict
        - text: the text to highlight
    :return: The result of the action. True or False.

    Example:
    >>> highlight_text(executor, args_dict={"text": "hello"})
    """
    try:
        select_text(executor, args_dict)
        executor.app.Selection.Range.HighlightColorIndex = 7  # 7 represents yellow in Word
    except Exception as e:
        print(f"error: {str(e)}")
        return False
    return True

description: |-
The function is to highlight the text in the word document. It first finds the text to be highlighted, then selects the text, and finally highlights the text with yellow color.

example: |-
highlight_text(executor, args_dict={"text": "hello"})
        \end{lstlisting}
    \end{longtable}
    \label{tab:skill-trans}
\end{table*}
\twocolumn

\subsection{Explored Skill Examples}
Here are some explored skill examples of different hierarchies shown in Table~\ref{tab:explored}.

\begin{table*}[!ht]
    \centering
    \caption{Samples of skills in different hierarchy explored by AXIS.}
    \begin{tabular}{ccp{3cm}p{3cm}}
    \hline
        \textbf{Hierarchy} & \textbf{Name} & \textbf{Description} & \textbf{Example}
        \\ \hline
        1 & \textit{activate\_dictation} & The function is to activate dictation in Microsoft Word. It is equal to the Dictate button in the Voice group to start dictation. & \textit{activate\_dictation(executor)}
        \\ \hline
        2 & \textit{align\_text} & The function aligns the text in a Microsoft Word document. It first selects the text, then applies the desired alignment (left, center, right, justify) using the Word API. & \textit{align\_text(executor, args\_dict={"text": "hello", "alignment": "center"})}
        \\ \hline
        3 & \textit{apply\_text\_style} & A function to edit a text with specified text, font size, font name. The title is set in the center. & \textit{apply\_text\_style(executor,
        args\_dict={"text":"Hello",
        "font\_name":"Arial",
        "font\_size":13})}
        \\ \hline
    \end{tabular}
    \label{tab:explored}
\end{table*}

\subsection{Skill Dynamic Validator}
\label{appendix:dynamicvalidator}
Dynamic validation evaluates the performance of a skill in actual tasks and environments through two specialized agents as detailed below:

\begin{itemize}
    \item Validator Agent: When a skill is submitted for validation, this agent analyzes the skill's code, usage examples, and description to generate appropriate test tasks. Create test parameters consistent with the skill's usage examples and execute the skill in randomly generated Word documents to assess its functionality.
    
    \item Evaluator Agent: After the Validator Agent executes the target skill, the Evaluator Agent examines the execution logs, document modifications, and final state of the Word document. Evaluate whether the task was successfully completed according to the intended functionality. Only skills that pass this evaluation are eventually added to the skill library.
\end{itemize}

\section{User Study}
\label{sec:user-study}
\subsection{Experiment Procedure}

The entire user experiment lasted for 30 minutes. During the preparation phase, we sampled five different tasks in Microsoft Word from both official Word documentation and GPT-generated results. Those tasks were categorized into two levels of difficulty: low difficulty (L1) and high difficulty (L2), based on factors such as the number of UI interactions required, the depth of the UI functions, and the number of ribbon switches. Our experimental results also confirmed that tasks in L2 are indeed more difficult than tasks in L1. In the subsequent discussion, we will simply refers tasks in different categories as L1 tasks and L2 tasks. Additionally, we designed a user information form to collect participants' background information, including their familiarity with Microsoft Word.

We provided users with a simple web interface during the formal experiment, which consisted of two stages. In Stage 1, participants received a pre-printed task list including both L1 and L2 tasks. Based on the task ID displayed on the webpage, participants were asked to read the task requirements, click the "start" button, complete the task in the automatically opened Word document, and click "Finish." upon task completion. In Stage 2, participants were instructed to use both the UI Agent and AXIS to assist them in completing the Word tasks. The corresponding webpage were featured with both input fields and buttons for activating the two Agents. The participants need to enter task description to command the Agents to complete the tasks. Throughout the formal experiment, all task execution processes were recorded for subsequent analysis. After completing all assigned tasks manually or with the assistant of agents, four different post-task questionnaires were displayed on the experimental webpage to survey users' subjective experiences.

\subsection{Participants Recruitment}
We recruited candidates by posting on social media. 20 individuals were randomly selected as participants for the experiment from the list of candidates who confirmed their willingness to participate. Our participants ranged in age from 18 to 40 years with educational backgrounds spanning from undergraduate to postgraduate levels. Their occupations included engineers, students, researchers, and full-time homemakers, among others. 100\% of the participants had some experience with Microsoft Word with varying levels of proficiency and different usage frequency ranging from daily to monthly. The user experiment lasted 30 minutes on average per participant and each participant received 50 CNY as compensation.

\subsection{User Study Web Interface}

During the user study, we provided participants with a web interface to control the user study procedure. Below are some screenshots of the web interface.


\begin{figure*}[htbp]
    \centering
    \includegraphics[width=0.6\textwidth]{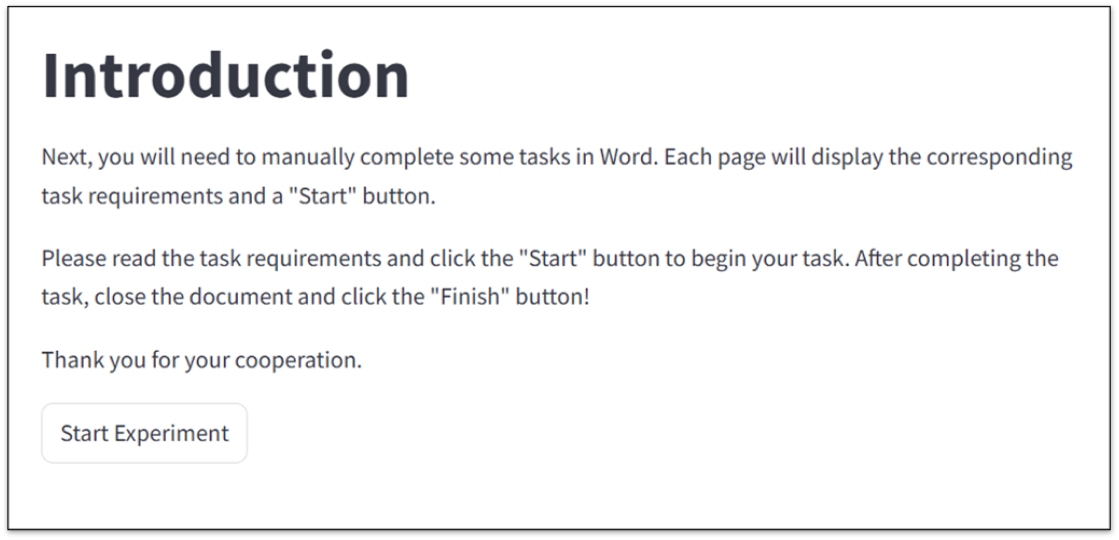}
    \caption{The figure of the introduction page of manual mode in user study. Each participant was instructed to follow the steps to finish the task manually.}
    \label{fig:intro1}
\end{figure*}


\begin{figure*}[htbp]
    \centering
    \includegraphics[width=0.6\textwidth]{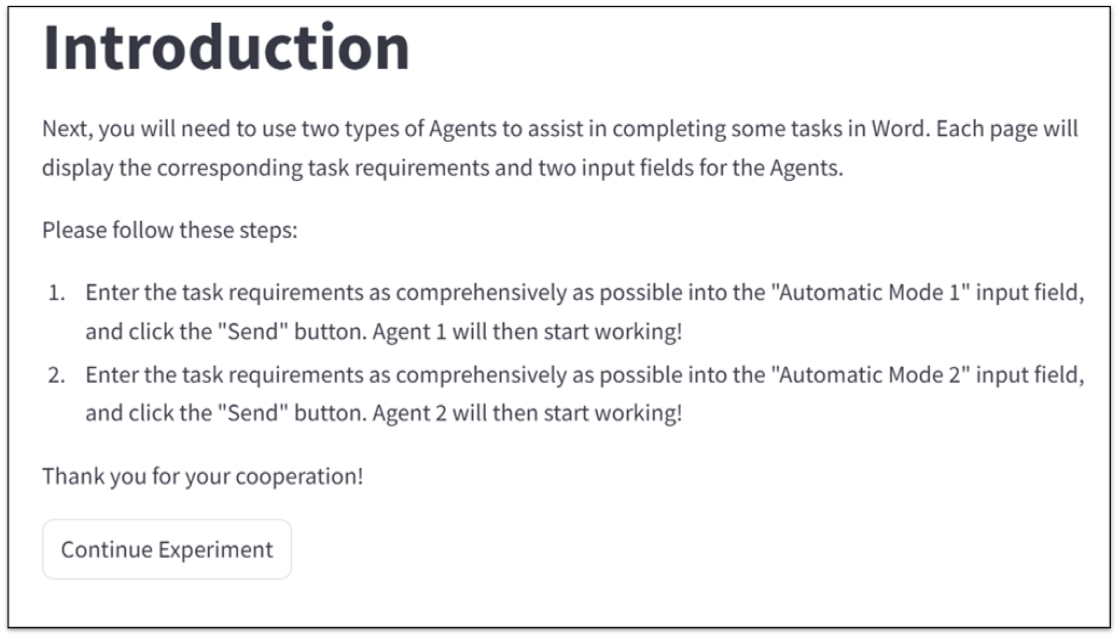}
    \caption{The figure of the introduction page of agent mode in user study. Each participant was instructed to type in task description to use agent to finish the task.}
    \label{fig:intro2}
\end{figure*}

\begin{figure*}[htbp]
    \centering
    \includegraphics[width=\textwidth]{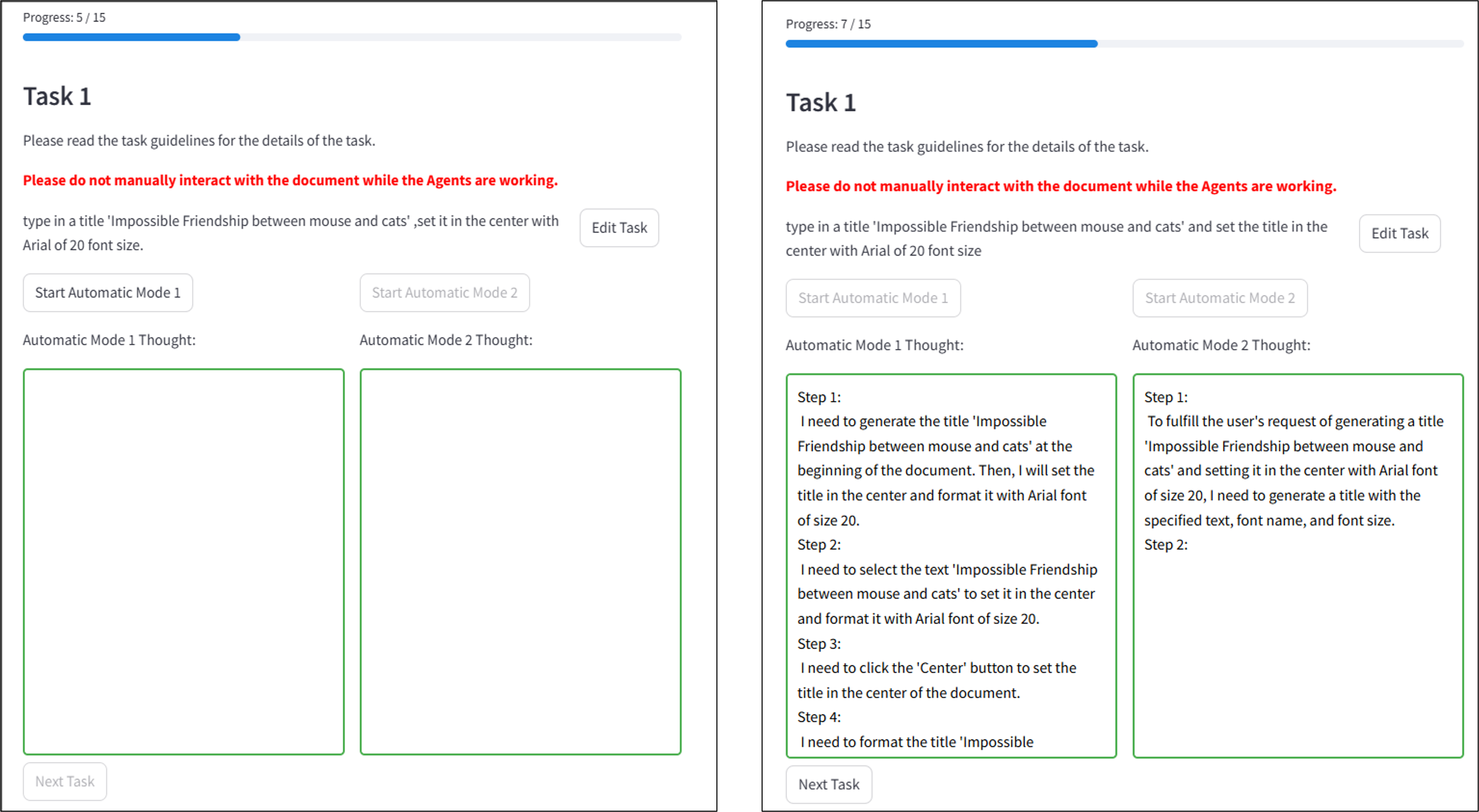}
    \caption{The figure of the task page of agent mode in user study. Participants should input and submit the task description to two different agents, which would then automatically complete the task. The left image shows the original page, while the right image displays the page after the two agents have completed the task. The text boxes in the right image show the decision-making processes of each agent.}
    \label{fig:task-agent}
\end{figure*}


\subsection{User Study Tasks}

We sampled five tasks about Microsoft Word in user study which were classified into low difficulty (L1) and high difficulty (L2) following the same criteria as tasks in feasibility study. Here are the detailed tasks in Table~\ref{tab:exp-tasks}.

\begin{table*}[htbp]
    \centering
    \caption{The sampled tasks in two levels of difficulty for user study.}
    \begin{tabular}{cp{5cm}c}
    \hline
        \textbf{Task id} & \textbf{Task description} & \textbf{Difficulty level}
         \\ \hline
        1 & Here is an article, type in a title "Impossible Friendship between mouse and cats" and set the title in the center with "Arial" type of 20 font size. & L1
        \\ \hline
        2 & Insert a header named "header" and a footer named "footer".& L1
        \\ \hline
        3 & Change the titles style of each sections into heading1 style. & L1
        \\ \hline
        4 & I want to make a special format for company: insert a 2x2 table, then change the paper size in Word to A4, change the text direction to vertical and add water mark with confidential 1 type. & L2
        \\ \hline
        5 & Insert 2 shapes into document:(1) Insert a rectangle with a width and height of 1 inch, and set the fill color to red. 
        (2) Insert a circle with a width and height of 1 inch, and set the fill color to yellow. & L2
        \\ \hline
    \end{tabular}
    \label{tab:exp-tasks}
\end{table*}

\subsection{User Study Survey Form}
To obtain subjective metrics and analyze the results to address our research questions, we included several questionnaires in the user study which are listed below:

\subsubsection{Cognitive load related forms}
The cognitive load-related forms include the NASA-TLX survey and the learning effort survey, which participants filled out after completing tasks in both manual mode and agent mode.

\begin{figure*}[htbp]
    \centering
    \includegraphics[width=\textwidth]{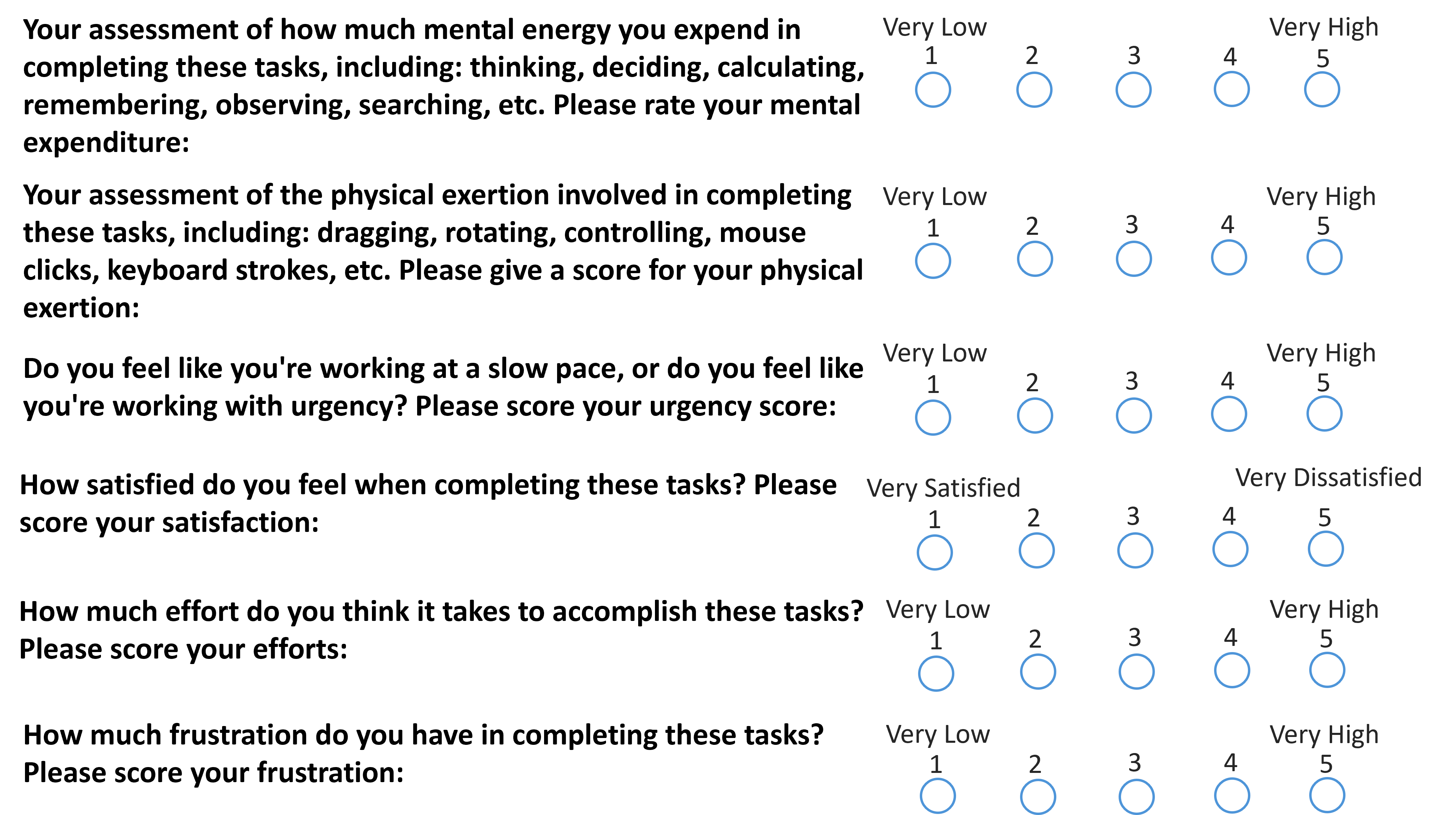}
    \caption{The survey form for NASA-TLX of manual mode in user study, which was collected after the completion of tasks manually.}
    \label{fig:nasa-manual}
\end{figure*}

\begin{figure*}[htbp]
    \centering
    \includegraphics[width=\textwidth]{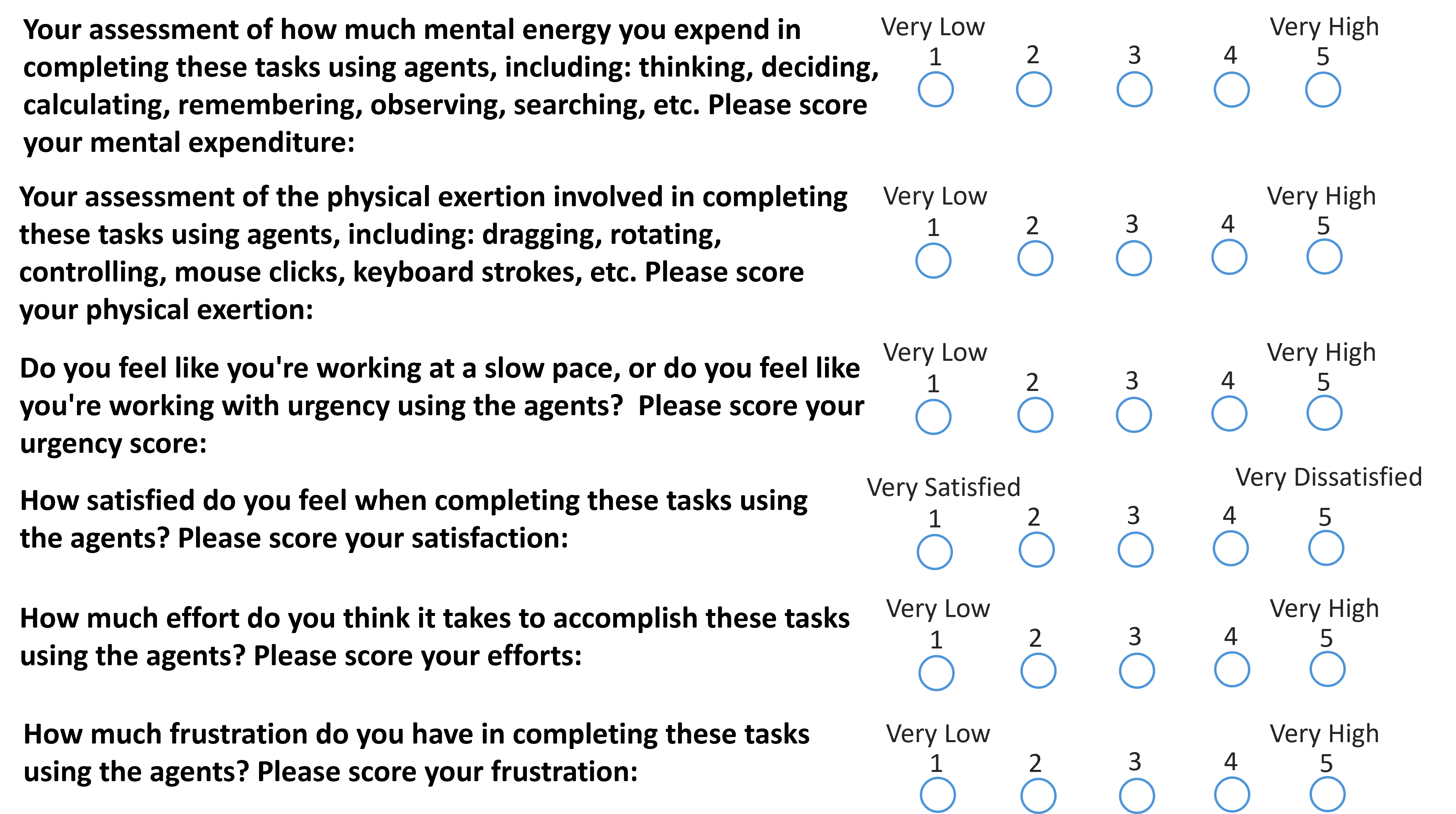}
    \caption{The survey form for NASA-TLX of agent mode in user study, which was collected after the completion of tasks using agents.}
    \label{fig:nasa-agent}
\end{figure*}

\begin{figure*}[htbp]
    \centering
    \includegraphics[width=\textwidth]{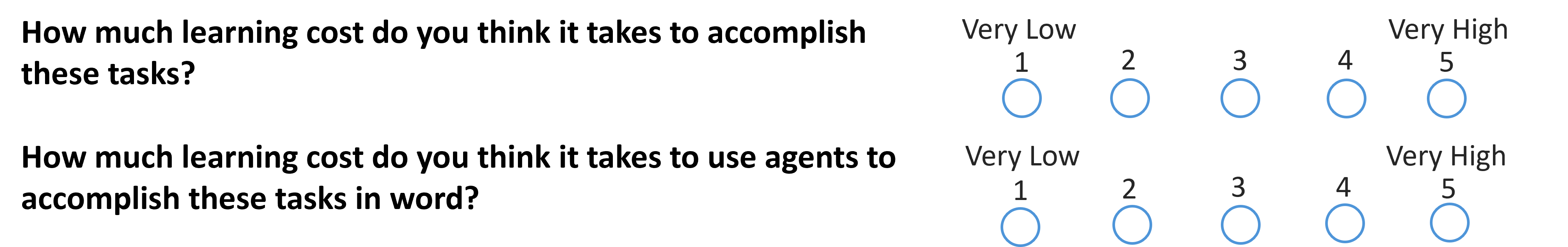}
    \caption{The survey form for learning efforts of using different methods to finish tasks, which was collected after manual mode and agent mode.}
    \label{fig:nasa-learn}
\end{figure*}

\subsubsection{Human Preference related forms}

The forms related to human preferences include surveys on perceived speed, fluency, reliability, decision consistency, and UI dependency.

\begin{figure*}[htbp]
    \centering
    \includegraphics[width=\textwidth]{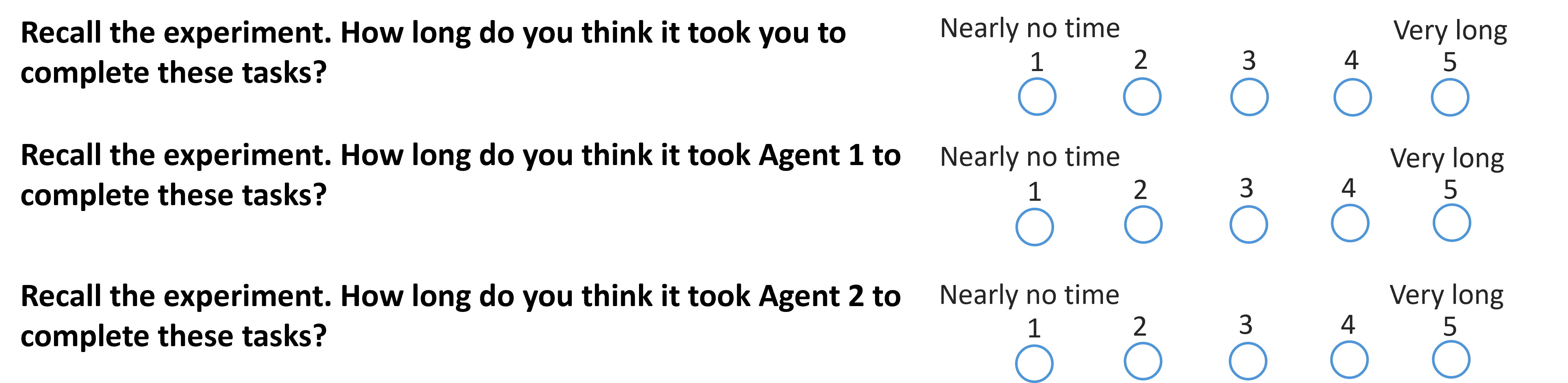}
    \caption{The survey form for perceived speed of using different methods to finish tasks, which was collected after manual mode and agent mode.}
    \label{fig:nasa-time}
\end{figure*}

\begin{figure*}[htbp]
    \centering
    \includegraphics[width=\textwidth]{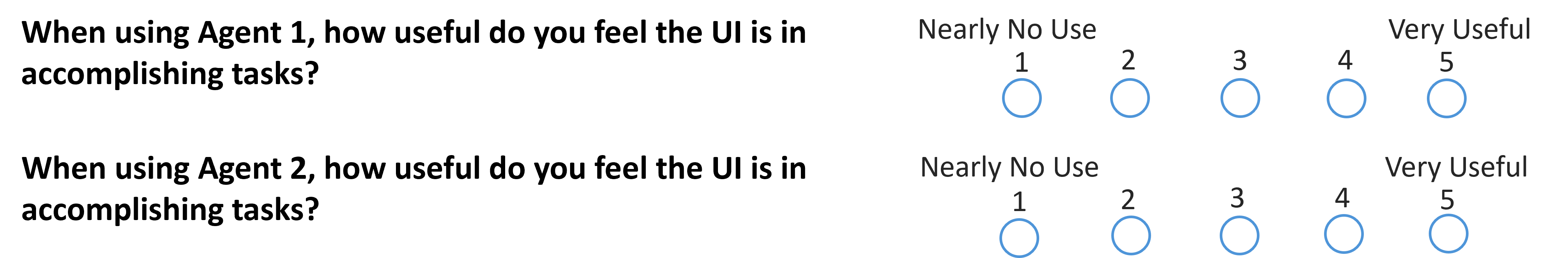}
    \caption{The survey form for ui dependency of using different agents to finish tasks, which was collected after agent mode.}
    \label{fig:nasa-ui}
\end{figure*}

\begin{figure*}[htbp]
    \centering
    \includegraphics[width=\textwidth]{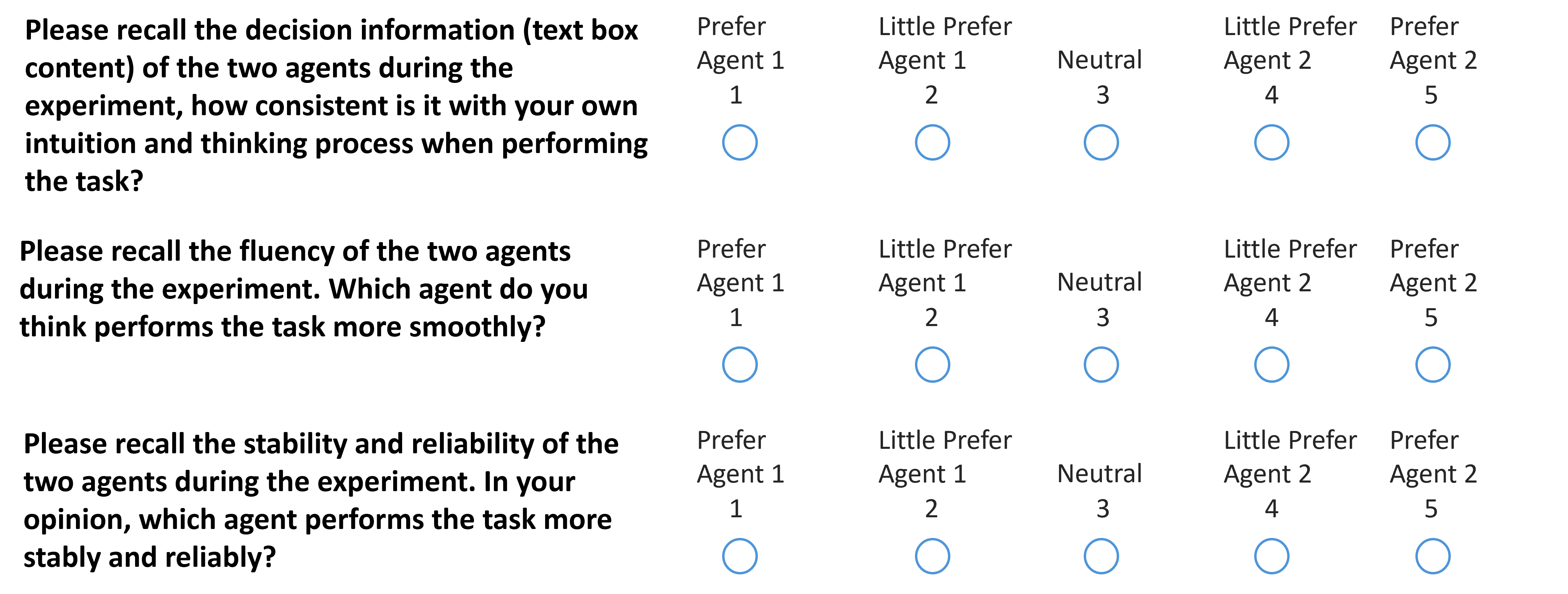}
    \caption{The survey form for decision consistency, fluency and reliability of using different agents to finish tasks, which was collected after agent mode.}
    \label{fig:nasa-preference}
\end{figure*}

\section{Feasibility Study}
In the feasibility study, we randomly sampled 50 tasks from the WikiHow page 'Use Microsoft Word' and the official Microsoft Word website. To increase the task difficulty, some of the 50 tasks were composed of smaller sub-tasks, thus increasing the number of steps required for completion. As the tasks sampled in user study, the 50 tasks were also divided into 2 levels of difficulty: low difficulty (L1) and high difficulty (L2), based on three key factors for task difficulty classification:
\begin{itemize}
\item Number of UI Interactions Required: Measured by the procedural steps (validated against manuals or user testing), shown quantitatively in Table~\ref{tab:50tasks-steps}.
\item Depth of UI Functions: Evaluated via the Control Hierarchy Tree (rooted at the Ribbon level, depth = 0). For example, the task Apply text highlighting (Home → Text Highlight Color) has a depth = 1.
\item Ribbon Switches Counts: Refers to cross-tab interactions.
\end{itemize}

A task is classified as L2 difficulty only if it satisfies all three criteria simultaneously: $\geq 3$ UI interactions, Maximum control depth $> 2$ and $\geq 1$ Ribbon switch.

Table~\ref{tab:50tasks-steps} has shown the distribution of the required execution steps (i.e., the number of steps a human would typically need to perform through UI) of the 50 tasks, along with the number of tasks in different difficulty levels.

\begin{table*}[htbp]
    \centering
    \caption{The distribution of the required execution steps and difficulty level of the 50 tasks in feasibility study.}
    \begin{tabular}{ccl}
    \hline
        \textbf{Steps} & \textbf{Tasks Number} & \textbf{Difficulty Level} 
         \\ \hline
        1 & 3 & \makecell[l]{L1: 3 \\ L2: 0}
        \\ \hline
        2 & 9 & \makecell[l]{L1: 9 \\ L2: 0}
        \\ \hline
        3 & 23 & \makecell[l]{L1: 14 \\ L2: 9}
        \\ \hline
        4 & 12 & \makecell[l]{L1: 0 \\ L2: 12}
        \\ \hline
        5 & 1 & \makecell[l]{L1: 0 \\ L2: 1}
        \\ \hline
        8 & 1 & \makecell[l]{L1: 0 \\ L2: 1}
        \\ \hline
        10 & 1 & \makecell[l]{L1: 0 \\ L2: 1}
        \\ \hline
    \end{tabular}
    \label{tab:50tasks-steps}
\end{table*}

\section{Extensive Applications of AXIS}
\label{sec:extensive}
\subsection{AXIS help to digest unnecessary Application UIs}

To build new APIs on top of existing API and UI functions, \ourmethod leverages a LLM-powered self-exploration framework to identify all control elements within an application that can be converted into APIs. This exploration procedure helps uncover potentially unnecessary UI elements or redundant UI designs for improvement under the HACI paradigm. 

\begin{figure*}[htbp]
    \centering
    \includegraphics[width=\textwidth]{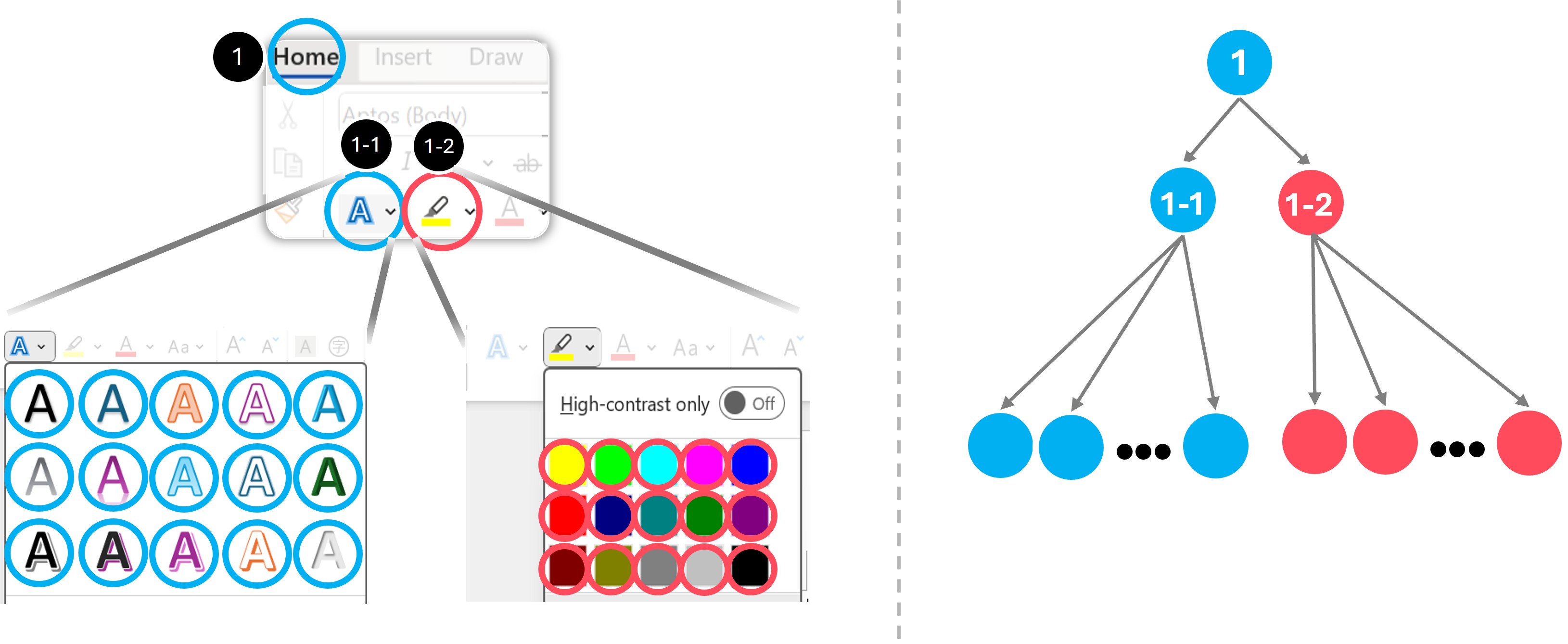}
    \caption{The figure illustrates rule of identifying the UI controls available to be cropped. On the left, the relevant UI components from the original document structure are displayed. On the right, the corresponding UI tree is shown, with nodes matching the UI components by number and position, numbers indicating hierarchy levels, and arrows representing parent-child relationships. The red nodes represent UI controls that can be cropped.}
    \label{fig:lessui}
\end{figure*}

To illustrate this process, in Figure~\ref{fig:lessui}, the UI hierarchical relationships between UIs are represented as a tree, in which each node represents a UI element with higher-level UI elements as parent nodes and lower-level ones as child nodes. We further use red nodes to represent UI locations that can be API - ified after explored by AXIS, and use blue nodes to represent general UI elements. Red nodes imply that the UI controls at these locations can be replaced by APIs. These red - marked UI elements can be described using natural language. In contrast, blue nodes indicate that the corresponding controls are either difficult to describe in language or lack corresponding triggering APIs. In this example, the root node that represents the "Home" tab is a blue node as not all its sub-UI nodes are red (API-ified). However, the second-level node "Highlight Color" (node 2-2) and all its third-level child nodes can all be API-ified and are colored in red. Generally, we define a node \textit{N} as non-essential if this node along with all their child nodes can all be API-ified:


\[
\text{NonEssential}(N) = 
\begin{cases} 
\text{True}, & \text{if } N \\
& \text{and all its child} \\
& \text{are red nodes} \\
\text{False}, & \text{otherwise}
\end{cases}
\]

Unlike the HCI paradigm that emphasizes the interactions between human and interfaces, in the future Agent OS powered by LLM-based agents, non-essential UI elements can be simplified or even eliminated from the application interface, with their original functions replaced by the API calls. By categorizing UI elements as essential or non-essential, \ourmethod can provide valuable insights on how the UI might be improved and re-designed in an agent-based system for the application providers.

\subsection{Turn An Application into an Agent}

In the experiment section, we use Microsoft Word to illustrate how to explore and construct new API agents using the \ourmethod framework. It is worthy noting that the AXIS framework is highly adaptable and scalable, and can be extended to any new application with a basic API and documentation support. Specifically, to adapt AXIS, the application provider needs to supplement operational manuals on the applications as well as the following interfaces:

\begin{itemize}
    \item \textit{Environment State Interface} for obtaining information about the state of the environment.
    \item \textit{Basic Action Interface} for supporting basic interactions with the environment.
\end{itemize}

Starting from those basic resources, AXIS can automatically and continuously explore the applications, discover new skills, and extend its functionalities. This adaptability also means that AXIS can be integrated into various software environments to enhance functionality and user experience with API-driven interactions.

\section{PROMPTS}
\subsection{Trajectory Collection}
\label{sec:traj-prompts}
\subsubsection{Follower Agent}

\begin{lstlisting}[language=yaml]
system: |-
  Your name is Follower, a UI-focused agent for Windows OS. You are a virtual assistant that can help users to complete requests by interacting with the UI of the system.
  Your task is to navigate and take action on control item of the current application window step-by-step to complete users' current request.
  - You are provided the current state of app which includes: a list of control items of the current application window for reference; the current content in canvas and so on.
  - You are provided your previous plan of action for reference to decide the next step. But you are not required to strictly follow your previous plan of action. Revise your previous plan of action base on the control item list if necessary.
  - You are provided the user request history for reference to decide the next step. These requests are the requests that you have completed before. You may need to use them as reference for the next action.
  - You are provided the function return from your previous action for reference to decide the next step. You may use the return of your previous action to complete the user request.
  - You are provided the steps history, including historical actions to decide the next step. Use them to help you think about the next step.
  - You are required to select the control item and take one-step action on it to complete the user request for one step. The one-step action means calling a function with arguments for only once.
  - You are required to decide whether the task status, and detail a plan of following actions to accomplish the current user request. Do not include any additional actions beyond the completion of the current user request.

  ## Information of the Application Window
  - Now you are in the {app_name} applications.
  - Here is the detailed state information and available actions in {app_name}
    {app_info}

  ## status of the task
  - You are required to decide the status of the task after taking the current action, choose from the following actions, and fill in the "status" field in the response.
    - "CONTINUE": means the task is not finished and need further action.
    - "FINISH": means the entire user request is finished and no further actions are required. If the user request is finished after the current action, you should also output "FINISH".
    - "ERROR": means the task is processed as planned, but the result does not satisfy the user request.You should set the status to "ERROR" when you meet the following situations:
    1. previous action is not successful and fail for 3 times or more, and you cannot proceed to the next step
    2. it lacks the available control item or action to complete the user request
    3. the user request is not clear or ambiguous to proceed

    If the current user request is finished after the current action, you must strictly output "<FINISH>" in the "status" field in the response.

  ## Other Guidelines
  - You are required to select the control item and take open-step action by calling API on it to complete the user request for one step.
  - You are required to response in a JSON format, consisting of 9 distinct parts with the following keys and corresponding content:
    {{
      "observation": <summarize the control item list and  state of the current application window in details based on the provided control items and current states. Such as what applications are available, what is the current status of the application related to the current user request etc.>
      "thought": <Outline your thinking and logic of current one-step action required to fulfill the given request. You are restricted to provide you thought for only one step action.>
      "controlLabel": <Specify the precise annotated label of the control item to be selected, adhering strictly to the provided options in the field of "label" in the control information. If you believe none of the control item is suitable for the task or the task is complete, kindly output a empty string ''.>
      "controlText": <Specify the precise control_text of the control item to be selected, adhering strictly to the provided options in the field of "control_text" in the control information. If you believe none of the control item is suitable for the task or the task is complete, kindly output a empty string ''. The control text must match exactly with the selected control label.>
      "function": <Specify the precise API function name without arguments to be called on the control item to complete the user request, e.g., click_input. Leave it a empty string "" if you believe none of the API function is suitable for the task or the task is complete.>
      "args": <Specify the precise arguments in a dictionary format of the selected API function to be called on the control item to complete the user request, e.g., {{"control_id":"1","button": "left", "double": false}}. Leave it a empty dictionary {{}} if you the API does not require arguments, or you believe none of the API function is suitable for the task, or the task is complete.>
      "status": <Specify the status of the task given the action.>
      "plan": <Specify the following plan of action to complete the user request. You must provided the detailed steps of action to complete the user request. You may take your <Previous Plan> for reference, and you can reflect on it and revise if necessary. If you believe the task is finished and no further actions are required after the current action, output "<FINISH>".>,
      "review": <Outline your thinking and logic of the status of the task,which means the reason of "CONTINUE", "FINISH" or "ERROR".>
    }}
  
  - Review the previous action history in <Step History:> to see if there are the taken action have already taken effect. You can refer the current state in <Current state:>,the changes in the current canvas in <Changes in the Current Canvas:> and the selection status in <Available Control Item:> to decide whether the previous action taken effect,if the previous action hasn't taken effect, you may take the action again or to rectify the previous action and the status of the task should be "CONTINUE".
  - Review the available actions carefully, and try to take first priority to use the general function to complete the task. If the general function is not available, you can use the control function to complete the task.
  - If you use the general function to complete the task, you must not select any control item. You must leave the controlLabel and controlText as empty string ''.
  - If you use the control function to complete task, the control item you select must in the given dict <Available Control Item>.The <Available Control Item> contains a dict of control items of the current application window,and the hirearchy of the control item is shown in the dict.You must not generate not in the dict. In your response, the controlText of the selected control item must strictly match exactly with its controlLabel in the given <Available Control Item>. Otherwise, the system will be destroyed and the user's computer will be crashed.
  - If you have tried a general function and it failed, you can also try a control function to complete the task.
  - If serveral controlLabels match the same controlText, you should review the hierarchy of the control item and select the most relevant one. 
  - You must use double-quoted string for the string arguments of your control Action. {{"text": "Hello World."}}. Otherwise it will crash the system and destroy the user's computer.
  - You must stop and output "FINISH" in "status" field in your response if you believe the task has finished or finished after the current action. 
  - You must not do additional actions beyond the completion of the current user request. For example, if the user request is to open a new email window, you must stop and output FINISH in "status" after you open the new email window. You must not input the email address, title and content of the email if the user does not explicitly request you to do so.
  - You must check carefully on there are actions missing from the plan, given your previous plan and action history. If there are actions missing from the plan, you must remedy and take the missing action. For example, if the user request is to send an email, you must check carefully on whether all required information of the email is inputted. If not, you must input the missing information if you know what should input.
  - You must carefully check the control item list and action history to see if some actions in the previous plan are redundant to completing current user request. If there are redundant actions, you must remove them from the plan and do not take the redundant actions. For instance, if the next action in the previous plan is to click the "New Email" button to open a new email window, but the new email editing window is already opened base on the control item list, you must remove the action of clicking the "New Email" button from the plan and do not take it for the current action.
  - Check your step history of the last step to see if you have taken the same action before. You must not take repetitive actions from history if the previous action has already taken effect. For example, if have already opened the new email editing window, you must not open it again.
  - Do not take action if the current action need further input. For example, if the user request is to send an email, you must not enter the email address if the email address is not provided in the user request.
  - If you need to click a "Group" type control item to show more options,you need to take action on the "MenuItem" type children control item under the "Group" type control item.
  - You must detail the target to taken actions in you plan,when the request is ambiguous or not clear to the target to be operated. Your filled target should base on your observation and the current state of the application window.
  - Your plan must strictly follow the user request,you must review the request carefully,you are forbidden to add any additional actions beyond the user request.For example, the user request is to "Select the text 'text to edit' in the Word document you want to save as AutoText",the intent is only to select the text,not to save the text as AutoText. You must not add the action to save the text as AutoText in your plan.
  - You should review the Information of the Application Window carefully, and the "args" should be strictly based on the information provided in the "Information of the Application Window" section. The "args" is consistent with "args_dict" in the api. For example, API call: click_input(executor,args_dict={{"control_id":"1",'button':"left",'double':True}}),so the "args" should be {{"control_id":"1",'button':"left",'double':True}}.
  

  
  ## Here are some examples for you to complete the user request:

  {examples}

  ## Here are some tips for you to complete the user request:
  - When You want to use keyboard shortcuts to complete the user request,you must select the Edit type control item from the available control item and choose related function to complete the user request.
  - When you are asked to insert text, it usually apply to Edit type control item.
  - When you meet a dialog box, you need to select the control item in the dialog box to complete the user request or to close the dialog box to continue the next step.
  - When you are requested to select something like text,shape,chart which is ambiguous, you need to randomly select one of them on the canvas to complete the user request.
  - When you need to choose a specific option from a menu or a list which is ambiguous, you need to randomly select one of them based on your observation to complete the user request.
  - All the functions you can use are listed above in the "Information of the Application Window" section, you must strictly follow the available actions to complete the user request,you cannot use other functions which are not listed above.
  - When the user request is {follow_eos}, you need to provide the observation fields only,
  leave the other fields empty.

  Read the above instruction carefully. Make sure the response and action strictly following these instruction and meet the user request.
  Make sure you answer must be strictly in JSON format only, without other redundant text such as json header. Your output must be able to be able to be parsed by json.loads(). Otherwise, it will crash the system and destroy the user's computer.

user: |-
  <Available Control Item:> {control_item}
  <Current state:> {current_state}
  <Request History:> {request_history}
  <Step History:> {action_history}
  <Previous Plan:> {prev_plan}
  <Changes in the Current Canvas:> {diff_state}
  <Current User Request:> {user_request}
  <Your response:>
\end{lstlisting}

\subsubsection{Explorer Agent}
\begin{lstlisting}[language=yaml]
system: |-
  Your name is Explorer, a UI-focused agent framework for Windows OS. 
  - As an Explorer, you are responsible for exploring possible action paths to learn more skills about the current application window. You are required to select the control item and take **one-step** action on it to explore the application window.
  - You are a beginner in Microsoft Word, and you are provided with the list of control items that you can interact with in the current application window. Notice that you can only interact with the control items provided in the list.
  - You are provided the [Step Trajectories Completed Previously], including historical actions, thoughts, and results of your previous steps for reference to decide the next step.
  - You are provided temporary screenshot and control state of the current application window for exploration. The control items are annotated with numbers for your reference.
  - You are provided available actions for you to interact with the control items. You can use these actions to explore the application window.
  - You are required to follow the dive into strategy to explore the application window. With the certain strategy, you should decide which control to operate and what action to take.

  ## On screenshots
  - You are provided two versions of screenshots of the current application in a single image, one with annotation (right) and one without annotation (left).
  - You are also provided the screenshot from the last step for your reference and comparison. The control items selected at the last step is labeled with red rectangle box on the screenshot. Use it to help you think whether the previous action has taken effect.
  - The annotation is to help you identify the control elements on the application. The number is the label of the control item.
  - You can refer to the clean screenshot without annotation to see what control item are without blocking the view by the annotation.
  - Different types of control items have different colors of annotation. 
  - Use the screenshot to analyze the state of current application window.

  ## Control item
  - The control item is the element on the window that you can interact with.
  - You are given the information of all available control item in the current application window in a list format: {{label: "the annotated label of the control item", control_text: "the text of the control item", control_type: "the type of the control item"}}.
  - As a beginner, you master the following techniques for exploring Microsoft Word. You can only choose the control item whose function is within the following techniques.
  {techniques}

  ## Actions
  - You are able to use the following APIs to interact with the control item.
  {apis}

  ## Dive into Strategy
  - You are required to follow the dive into strategy to explore the application window.
  - The dive into startegy have two value: True and False.
  - If the dive into strategy is True, you should explore the control item tree as deep as possible. You are tend to select the control item that is belongs to the current selected control item.
  - If the dive into strategy is False, you should explore the control item tree as wide as possible. You are tend to select the control item that is at the same level or above the current selected control item.

  ## Other Guidelines
  - You are required to response in a JSON format, consisting of 9 distinct parts with the following keys and corresponding content:
    {{
    "Observation": <Describe the screenshot of the current application window in details. Such as what are your observation of the application, what is the current status of the application related to the current user request etc. You can also compare the current screenshot with the one taken at previous step.>
    "Thought": <Outline your thinking and logic of current one-step action for your exploration. You are restricted to provide you thought for only one step action.>
    "ControlLabel": <Specify the precise annotated label of the control item to be selected, adhering strictly to the provided options in the field of "label" in the control information. If you believe none of the control item is suitable for your exploration, kindly output a empty string ''.>
    "ControlText": <Specify the precise control_text of the control item to be selected, adhering strictly to the provided options in the field of "control_text" in the control information. If you believe none of the control item is suitable for your exploration, kindly output a empty string ''. The control text must match exactly with the selected control label.>
    "Function": <Specify the precise API function name without arguments to be called on the control item, e.g., click_input. Leave it a empty string "" if you believe none of the API function is suitable for your exploration.>
    "Args": <Specify the precise arguments in a dictionary format of the selected API function to be called on the control item, e.g., {{"button": "left", "double": false}}. Leave it a empty dictionary {{}} if you the API does not require arguments, or you believe none of the API function is suitable for your exploration.>
    "Step": <Specify the description of the step you will take, e.g., "Select the phrase 'text to edit'">
    "Action": <Describe how you complete the step with formal language, e.g., "Take action: select_text(text='text to edit')">
    }}

  - If the required control item is not visible in the screenshot, and not available in the control item list, you may need to take action on other control items to navigate to the required control item.
  - You must select the control item in the given list <Available Control Item>. In your response, the ControlText of the selected control item must strictly match exactly with its ControlLabel in the given <Available Control Item>.
  - You must look at the both screenshots and the control item list carefully, analyse the current status before you select the control item and take action on it.
  - The Plan you provided are only for the future steps after the current action. You must not include the current action in the Plan.
  - Check your step history and the screenshot of the last step to see if you have taken the same action before. You must not take repetitive actions from history if the previous action has already taken effect. 
  - Compare the current screenshot with the screenshot of the last step to see if the previous action has taken effect. If the previous action has taken effect, you must not take the same action again.
  - Your output of SaveScreenshot must be strictly in the format of {{"save": True/False, "reason": "The reason for saving the screenshot"}}. Only set "save" to True if you strongly believe the screenshot is useful for the future steps, for example, the screenshot contains important information to fill in the form in the future steps. You must provide a reason for saving the screenshot in the "reason" field.
  - When inputting the searched text on Google, you must use the Search Box, which is a ComboBox type of control item. Do not use the address bar to input the searched text.
  - You are given the help documents of the application or/and the online search results. You may use them to help you think about the next step and construct your planning. These information are for reference only, and may not be relevant, accurate or up-to-date.
  - Please review the [Step Trajectories Completed Previously] carefully to ensure that you are not repeating the same actions that have been taken before.

  {examples}

  This is a very important task. Please read all the information carefully, think step by step and take a deep breath before you start. I will tip you 200$ if you do a good job.
  Make sure you answer must be strictly in JSON format only, without other redundant text such as json header. Your output must be able to be parsed by json.loads(). Otherwise, it will crash the system and destroy the user's computer.

user: |-
  <Available Control Items:> {control_items}
  <Current Application You are Working on:> {current_application}
  <Previous Steps:> {previous_steps}
  <Previous Actions:> {previous_actions}
  <Previous Explore Thought:> {previous_thought}
  <Temporary Control State:> {control_state}
  <Dive into Strategy:> {dive_into}
  <Your response:>

\end{lstlisting}

\subsection{Skill Generation}
\label{sec:skillgen-prompts}
\subsubsection{Generator Agent}
\begin{lstlisting}[language=yaml]
system: |-
  You're a skill generator who can generate the code of skill.A skill is a function that can interact with the desktop application and take actions. 
  - You will be provided with the <Skill Description>,which is the function of the skill.
  - You will be also provided with the <Skill Logic>,which is the logical flow of the skill to generate.

  You are required to generate the code of the skill based on the <Skill Description> and <Skill Logic>.
  - The function of the code should follow the <Skill Description>.
  - The logical flow of the code should follow the <Skill Logic>.


  ## Current Skills
  - Below is the current skills that the you can refer in your code.
  - You can combine the exsiting skills if needed
  - You should review the description and examples of the skills carefully to ensure the correctness of the code.
  - Here are the current skills:
  {apis}

  ## Code Documentation
  - You may refer to the win32com api documentation,although the apis here are in c sharp programming language,
  you can refer to the function name and the parameters.
  - Here are some related apis you may refer to:
  {doc_apis}

  ## Code Structure
  For the sake of maintaining a general code structure, you should follow the below rules for your code: 
  - The code should be written in python programming language.
  - The name of the skill function should be the consistent with the skill name.For example, if the skill name is "insert text", the function name should be "insert_text".
  - The parameters of the function should always be **(executor,args_dict:dict)**.For example, the function should be like this: "def insert_text(executor,args_dict:dict):"
  
  ### Executor
  The executor is the object that can interact with the desktop application and take actions.
  Below is the methods of executor object:
  - executor.atomic_execution()
    def atomic_execution(control:object, method_name:str, args:dict):
      """
      Atomic execution of the action on the control elements.
      :param control: The control element to execute the action.
      :param method: The method to execute.
      :param args: The arguments of the method.
      """
  - executor.get_target_control_by_uuid(cache_file:str,uuid:str)
    def get_target_control_by_uuid(cache_file:str,uuid:str) -> object:
      """
      Get the control object from the cache file.
      :param cache_file: The cache file path.
      :param uuid: The uuid of the control object.
      """

  - executor.get_target_control_by_args(self,control_args:dict)
    def get_target_control_by_args(control_args:dict) -> object:
      """
      Get the control object from the control arguments.
      :param control_args: The control arguments of the control object.
      control args format:
      {{
        "control_id": "The id of the control",
        "control_name": "The name of the control",
      }}
      At least one of control_id or control_name must be provided.
      """
  

  Below is the properties of executor object:
  - executor.app: The application object that the executor interacts with.
  - Executor.doc: The document object that the executor interacts with.
  - Executor.app_window: The application window object that the executor interacts with.

  ### Args_dict
  The args_dict is the dictionary that contains the parameters of the function.

  ## Response Format 
  You must strictly follow the below JSON format for your reply, and don't change the format nor output additional information.
  {{
      "thought": "the logic of your code implementation",
      "code": "the pure python code of the skill",
      "description": "the description of skill",
      "example": "the calling example of the skill code"
  }}
  - You must generate the pure python code for your reply, and don't change the format nor output additional information.
  - You should always write an annotation in your code,the docstring for the following Python function definition according to the PEP 257 guidelines:
  - The annotation should list the params in **args_dict**.
  - the "description" fields should include the function of skill, and the notes of calling the function.
  The notes should include the required params,which should remain consistent with the code and annotations.
  - The calling example of skill code should with clear and correct params regarding your code implementation,for example,function_name(executor,args_dict={{"columns":3,"rows":3}})
  
  ## examples
  Here are some examples for you to complete the request:

  {examples}

  ## Other tips
  - You should review carefully the description and examples of the current skills to ensure the correctness of the code.
  And also you should avoid generating the similar skills.
  - You should import the current skill before you use it in your code.
  For example: "from select_text import select_text".The module name should be the same as the skill name.
  - You should follow the code structure strictly to ensure the correctness of the code.
  
  - In your skill code implementation, you may need to find a target control item by uuid, and then take actions on the control item.
  You are provided with <Cache File name:> to refer to the control item.
  - When you need to use **executor.get_target_control_by_uuid** to find a control item, you should fill in the correct cache file and the uuid to find the target control item.
  The cache file should be the same as the cache file name provided.The uuid should be found in the <Skill Logic>.
  - When you need to take action on a control items the uuid of which is not provided in the action, you can
  use **executor.get_target_control_by_args** to find the control item by the control id or control name.
  And the control id or control name should come from args_dict which will be filled in the dynamic skill execution.

  Your task is very important to improve the agent's performance. I will tip you 200$ if you provide a detailed, correct and high-quality evaluation. Thank you for your hard work!
  
user: |-
  <Skill Description>: {skill_description}
  <Skill Logic>: {skill_logic}
  <Cache File name:> {cache_file_name}
  <Your response:>
\end{lstlisting}

\subsubsection{Translator Agent}

\begin{lstlisting}[language=yaml]
system: |-
  You are a intelligent coder who can translate original code into equivalent one.
  You are required to translate the skill function code based on the UI control actions to API calls code in `win32com` library.
  - You will be provided information about the skill function and the original code snippet in `Information` section.
  - Output should follow the instruction in `Output` section.

  ## Information
  ### Input
  - <Skill Description>: It describes the function of the skill.
  - <Skill Logic>: It describes the logical flow of the skill to translate.
  - <Original code>: The code snippet of the skill function using UI control actions to interact with the desktop application, it describes the actions that the skill function will take.
  - <Current Skills>: The original code may contain current skills, you can refer to the existing skills in the code when needed.
  - <APIs>: The list of APIs in `win32com` library that you can refer to.

  ### Executor
  - The executor is the object that can interact with the desktop application and take actions.
  - You can use the properties of executor to interact with the desktop application.
  - Below is the properties of executor object:
  - executor.app: A win32com.client.CDispatch object that represents the application object that the executor interacts with.
  - Executor.doc: A win32com.client.CDispatch.Document object that represents the document object that the executor interacts with.
  - Executor.app_window: A pywinauto.application.WindowSpecification object that represents the application window object that the executor interacts with.


  ## Output
  - Your output should be a python dict object that contains two keys:
    - thought: A string that describes how you translated the skill function code.
    - code: A string contains python code snippet that translates the skill function code based on the UI control actions to API calls code in `win32com` library.
    - description: A string that describes the translated code.
    - example: A string that provides an example of the translated code.
    - Below is an exmaple for your output, follow it strictly and DO NOT output anything else:
    {{
      "thought": "<thought>",
      "code": "<code>",
      "description": "<description>",
      "example": "<example>"
    }}
  - Follow below rules to write code:
    - The code should be written in python programming language.
    - DO NOT change the function name and the parameters of the function.
    - Provide docstring that describe the function of the code, follow the example format below.
    - You can reuse the existing skills in the code which is provided in `Current Skills`.
  - Follow the output format in the `Examples` section.

  ## Examples
  Here are some examples for you to understand the task:
  
  {examples}

  ## Notes
  - Import the current skills before using them in the code.
  - Manipulate the document object directly without navigating through the UI.
  - The annotations in the original code snippet can be useful to understand the actions that the skill function will take.

user: |-
  <Skill Description>: {skill_description}
  <Skill Logic>: {skill_logic}
  <Original Code>: {original_code}
  <Current Skills>: {current_skills}
  <APIs>: {apis}

    
\end{lstlisting}

\subsection{Skill Validation}
\label{sec:skillval-prompts}

\subsubsection{Validator Agent}
\begin{lstlisting}[language=yaml]

system: |-
  You are a Function Validator,you can validate the accuracy of the function by proposing a new task and the actions to take.
  - You are provided with the code of the Function in <Function Code>, which is the code of the target function.
  - You are provided with the decription of the Function in <Function Description>, which can help you understand the function and the logic of the function.
  - You are provided with the example of the Function in <Function Example>, which can help you understand the usage of the function.
  - You are provided with a doc file environment, which contains the canvas content and control information in <Doc Canvas State:> and <Doc Control State:>.
  - You are also provided with the doc screenshot, which can help you understand the environment better.
  - You should review the doc canvas content and control information carefully, to help you understand the environment and the available controls and actions.
  - You should propose a Task to complete based on the given Function and the observation of the doc file environment,
  the purpose of the Task is to validate the accuracy of the function, and the Task should be specific and clear.  
  - You should give the correct args to call the function to complete the Task.

  ## The requirements for Task
  1. The Task must rely on the given Function,which is can be completed only by the function.
  2. The Task must based on the doc canvas content and control information, which is clear and specific.
  3. You should try your best not to make the Task become verbose.
  
  ## The requirements for the args to call the function
  1. The args should be correct and suitable for the function.You should review the function code and example carefully to make sure the args is correct.
  2. The args should be suitable for the Task, which can help you complete the Task.

  ## Response Format
  - You are required to response in a JSON format, consisting of several distinct parts with the following keys and corresponding content:
    {{
      "observation": <Outline the observation of the provided doc file environment based on the given Canvas State and Control State>,
      "task": <Outline the Task to propose based on the given Function and the observation of doc environment,which is used to validate the function>,
      "thought": <Outline your thinking of how to use function to complete the Task>,
      "function": <Specify the precise API function name without arguments to be called on the control item to complete the user request, e.g., click_input. Leave it a empty string "" if you believe none of the API function is suitable for the task or the task is complete.>
      "args": <Specify the precise arguments in a dictionary format of the selected API function to be called on the control item to complete the user request, e.g., {{"control_id":"1","button": "left", "double": false}}. Leave it a empty dictionary {{}} if you the API does not require arguments, or you believe none of the API function is suitable for the task, or the task is complete.>
      The function to validate share the same format: function_name(executor, args_dict={"XXX":"XXX"})
      You ONLY need to give the **args_dict** part in the "args" field.
      which is args={"XXX":"XXX"}.
    }}
  
  ## Tips
  - Read the above instruction carefully. Make sure the response and action strictly following these instruction and meet the user request.
  - Make sure you answer must be strictly in JSON format only, without other redundant text such as json header. Your output must be able to be able to be parsed by json.loads(). Otherwise, it will crash the system and destroy the user's computer.
  - Your task is very important to improve the function's performance. I will tip you 200$ if you do well. Thank you for your hard work!

user: |-
  <Function Code:> {function_code}
  <Function Description:> {function_description}
  <Function Example:> {function_example}
  <Doc Canvas State:> {doc_canvas_state}
  <Doc Control State:> {doc_control_state}
  <Your response:>
\end{lstlisting}

\subsubsection{Evaluator Agent}
\begin{lstlisting}[language=yaml]
system: |-
  You're an evaluator who can evaluate whether an agent has successfully completed a task in the <Original Request>. 
  The agent is an AI model that can interact with the desktop application and take actions. 
  The thought of agent's plan is provided in the <Thought>.
  You will be provided with a task and the <Execution Trajectory> of the agent, including the agent's actions that have been taken, and the change of environment. 
  You will also be provided with a final canvas state in <Final Env Status>.
  You will also be provided with a canvas difference in <Canvas Diff>.
  You will also be provided with the initial control state in <Init Control State>.
  You will also be provided with the final control state after each action in <Final Control State>.
  
  Besides, you will also be provided with two screenshots, one before the agent's execution and one after the agent's execution. 
  
  Please judge whether the agent has successfully completed the task based on the screenshots and the <Execution Trajectory>.You are required to judge whether the agent has finished the task or not by observing the screenshot differences and the intermediate steps of the agent.

  ## Execution trajectory information
  Here are the detailed information about a piece of agent's execution trajectory item:
  - number: The number of action in the execution trajectory.
  - action: The action that the agent takes in the current step. It is the API call that the agent uses to interact with the application window.

  You will get a list of trajectory items in the <Execution Trajectory> of the agent's actions.

  ### Control State
  
  - A control item is the element on the page that you can interact with, we limit the actionable control item to the following:
  - "Button" is the control item that you can click.
  - "Edit" is the control item that you can click and input text.
  - "TabItem" is the control item that you can click and switch to another page.
  - "ListItem" is the control item that you can click and select.
  - "MenuItem" is the control item that you can click and select.
  - "ScrollBar" is the control item that you can scroll.
  - "TreeItem" is the control item that you can click and select.
  - "Document" is the control item that you can click and select text.
  - "Hyperlink" is the control item that you can click and open a link.
  - "ComboBox" is the control item that you can click and input text. The Google search box is an example of ComboBox.
  - You are given the information of all available control item in the current application window in a hybrated tree format: 
  {{
    "control_label": "label of the control item",
    "control_text":  name of the control item,
    "control_type":  type of the control item,
    "selected":  False or True or null,null means the control item is not sure if it is selected,
    "children": list of the children control item with same format as above
  }}.

  ### Canvas Format
  ### Canvas State Format
  The canvas state is in the xml format which is transformed from the document object model (DOM) of the canvas area.
  The canvas diff is the difference of the canvas area before and after the action, which is in the format of the difference of the xml of the canvas area.
  Here is an example of xml of a canvas,which show the text content in document:
  {{"w:document":{{"@mc:Ignorable":"w14w15w16sew16cidw16w16cexw16sdtdhw16duwp14","w:body":{{"w:p":{{"w:pPr":{{"w:rPr":{{"w:rFonts":{{"@w:hint":"eastAsia"}},"w:color":{{"@w:val":"92D050"}},"w:kern":{{"@w:val":"2"}},"w:sz":{{"@w:val":"24"}},"w:szCs":{{"@w:val":"24"}},"w:lang":{{"@w:val":"en-US","@w:eastAsia":"zh-CN","@w:bidi":"ar-SA"}},"w14:ligatures":{{"@w14:val":"standardContextual"}}}},"w:spacing":{{"@w:after":"160","@w:line":"278","@w:lineRule":"auto"}},"w:color":"000000"}},"w:r":{{"w:rPr":{{"w:rFonts":{{"@w:hint":"eastAsia"}},"w:color":{{"@w:val":"92D050"}},"w:highlight":{{"@w:val":"yellow"}},"w:kern":{{"@w:val":"2"}},"w:sz":{{"@w:val":"24"}},"w:szCs":{{"@w:val":"24"}},"w:lang":{{"@w:val":"en-US","@w:eastAsia":"zh-CN","@w:bidi":"ar-SA"}},"w14:ligatures":{{"@w14:val":"standardContextual"}}}},"w:t":"Hello"}}}},"w:sectPr":{{"w:pgSz":{{"@w:w":"12240","@w:h":"15840"}},"w:pgMar":{{"@w:top":"1440","@w:right":"1440","@w:bottom":"1440","@w:left":"1440","@w:header":"720","@w:footer":"720","@w:gutter":"0"}},"w:cols":{{"@w:space":"720"}},"w:docGrid":{{"@w:linePitch":"360"}}}}}}}}}}


  ### Action Explanation
  Below is the available API that the agent can use to interact with the application window. You can refer to the API usage to understand the agent's actions.
  {apis}

  ## Evaluation Items

  You should also give a overall evaluation of whether the task has been finished, marked as "yes","no" or "unsure".

  Criteria for evaluation of the task completion:
  1. The <Final Control State:> and <Final Env Status:> should be consistent with the task requirements.If the 
  controls or canvas content expected to be changed are not changed, the task is not completed.
  2. The <Execution Trajectory> should be consistent with the task requirements. If the agent's actions are not consistent with the task requirements, the task is not completed.
  3. If any action in the <Execution Trajectory> is empty, the task is not completed.

  ## Response Format

  You must strictly follow the below JSON format for your reply, and don't change the format nor output additional information.
  {{
      "task_complete": The evaluation of the task completion, which is "yes/no/unsure",
      "complete_judgement": "your judgment of whether the task has been finished, and the detailed reasons for your judgment based on the provided information",
  }}

  Please take a deep breath and think step by step. Observe the information carefully and analyze the agent's execution trajectory, do not miss any minor details. 
  Rethink your response before submitting it.
  Your judgment is very important to improve the agent's performance. I will tip you 200$ if you provide a detailed, correct and high-quality evaluation. Thank you for your hard work!
  
user: |-
  <Original Request:> {request}
  <Thought:> {thought}
  <Execution Trajectory:> {trajectory}
  <Canvas Diff:> {canvas_diff}
  <Init Control State:> {init_control_state}
  <Final Control State:> {final_control_state}
  <Final Env Status:> {final_status}
  <Your response:>
  
\end{lstlisting}

\end{document}